\documentclass{article}

\usepackage{arxiv}

\usepackage[utf8]{inputenc}
\usepackage[T1]{fontenc}
\usepackage{hyperref}
\usepackage{url}
\usepackage{booktabs}
\usepackage{amsfonts}
\usepackage{amsmath}
\usepackage{amssymb}
\usepackage{nicefrac}
\usepackage{graphicx}
\usepackage{natbib}
\usepackage{doi}
\usepackage{xcolor}
\usepackage{multirow}
\usepackage{array}
\usepackage{tabularx}
\usepackage{caption}
\usepackage{subcaption}
\usepackage{float}

\graphicspath{{Figures/}}

\hypersetup{
  colorlinks=true,
  linkcolor=blue!70!black,
  citecolor=blue!70!black,
  urlcolor=blue!70!black,
  pdftitle={SynHeart Capacity: Operationalization of Cognitive Capacity State
    from Wearable Physiological Signals},
  pdfsubject={cs.LG, eess.SP, cs.HC},
  pdfauthor={Yisak T, Henok A, Israel G},
  pdfkeywords={capacity theory, mental workload, physiological computing,
    deep learning, multi-task learning, cardiac signals, electrodermal activity}
}

\title{  Synheart Capacity: A Theory-Driven Physiological Representation of Cognitive Capacity Dynamics from Wearable Signals }

\renewcommand{\shorttitle}{Synheart Capacity: Cognitive Capacity State from Wearable Signals}

\author{
  Yisak Debele\thanks{Corresponding author.
    \texttt{yisak@synheart.ai}}$^{1}$
  \quad
  Henok Ademtew$^{1}$
  \quad
  Israel Goytom$^{1}$
  \\[0.5em]
  $^{1}$Synheart AI, Vancouver, Canada \\
}

\date{April 2026}

\begin{document}
\maketitle

\begin{abstract}
Human cognitive performance is constrained by limited mental resources, yet
continuous computational estimation of cognitive capacity dynamics remains an
open challenge. We propose a theory-driven multimodal learning framework that
models capacity-related cognitive state as a two-dimensional physiological
representation defined by voluntary resource allocation (mental effort) and
overload-related strain (stress). The proposed architecture combines dual-stream
encoding of cardiac (IBI/HRV) and electrodermal (EDA) signals with late fusion
and task-specific output heads that independently estimate probabilistic effort
and stress states.

Evaluation on the SWELL-KW dataset using strict leave-one-subject-out
cross-validation demonstrates cross-individual generalization (stress:
70.0\% balanced accuracy; effort: 72.2\%), with significant gains from
multimodal integration and theory-guided supervision. Rather than collapsing
physiological dynamics into a single workload label, the proposed effort--stress
state-space enables structured differentiation between distinct cognitive
regimes, including productive engagement and overload-related strain.
Predicted state trajectories exhibit significant demand-sensitive shifts under
controlled workload manipulations, with effort and stress responding
differentially across interruption and time-pressure conditions.

These results suggest that physiologically grounded multidimensional state
representations may provide a foundation for adaptive systems capable of
continuous capacity-aware monitoring and human-centered interaction.
\end{abstract}

\keywords{capacity theory \and mental workload \and physiological computing
  \and deep learning \and multi-task learning \and real-time estimation
  \and cardiac signals \and electrodermal activity \and cognitive load
  \and adaptive systems}

\section{Introduction}

\subsection{Background and Motivation}

Human performance failures---whether in classrooms, workplaces, or
safety-critical environments---often emerge when cognitive demands exceed
available mental resources. Capacity theory proposes that individuals possess
limited processing resources that must be dynamically allocated to task demands,
and performance deteriorates when demand approaches or exceeds available
capacity~\citep{kahneman1973,wickens2002}. Within this framework, voluntary
resource allocation (mental effort) and involuntary strain responses (stress)
represent two interacting mechanisms governing performance under
load~\citep{hockey1997,matthews2002}. In this work, we use the term
\emph{capacity state} to denote the dynamic configuration of effort (resource
allocation) and stress (overload pressure) that reflects an individual's
functional cognitive regime.

While capacity theory has guided decades of behavioral and neurocognitive
research, modern intelligent systems remain largely unaware of users' cognitive
limits. Adaptive tutoring platforms, decision-support tools, and
human--automation interfaces increasingly aim to personalize difficulty and
assistance~\citep{chen2008,parasuraman2000}. However, these systems typically
rely on behavioral metrics such as error rates or response times, detecting
overload only after performance has degraded~\citep{backs2000}. Without
continuous estimation of cognitive state, AI systems cannot distinguish between
productive engagement and emerging overload. To investigate this challenge, the
present work draws on the SWELL Knowledge Work dataset~\citep{koldijk2014}, a
publicly available corpus of physiological recordings from 25 participants
performing office tasks under controlled cognitive demand manipulations.

A central challenge lies in the fact that capacity is a latent construct: it
cannot be measured directly but must be inferred from allocation and strain
processes. Traditional operationalizations---dual-task paradigms, subjective
workload ratings (e.g., NASA-TLX), or neuroimaging---provide indirect and
discontinuous measures~\citep{gopher1986,rubio2004}. These methods lack
ecological validity and are unsuitable for real-time adaptive deployment.

Effort and stress are theoretically related but not equivalent constructs.
High effort may reflect adaptive engagement under manageable demand, whereas
high stress may indicate that task demands are approaching or exceeding
available cognitive resources~\citep{hockey1997,matthews2002}. As a result,
collapsing these processes into a single workload label obscures an important
distinction between sustainable engagement and overload-related strain. This
distinction is particularly important for adaptive human-centered systems,
which may require fundamentally different interventions depending on whether
a user is productively challenged or approaching cognitive overload.

Physiological signals offer a pathway toward continuous, objective inference of
cognitive dynamics. Cardiac variability reflects autonomic regulation associated
with executive control and effort mobilization~\citep{mulder1992,thayer2009},
while electrodermal activity indexes sympathetic arousal linked to stress and
strain~\citep{boucsein2012,critchley2002}. With the proliferation of wearable
sensors and edge-capable machine learning, these signals can now be acquired and
processed unobtrusively in real-world contexts~\citep{poh2010,can2019}.

\begin{figure}[htbp]
  \centering
  \includegraphics[width=0.85\linewidth]{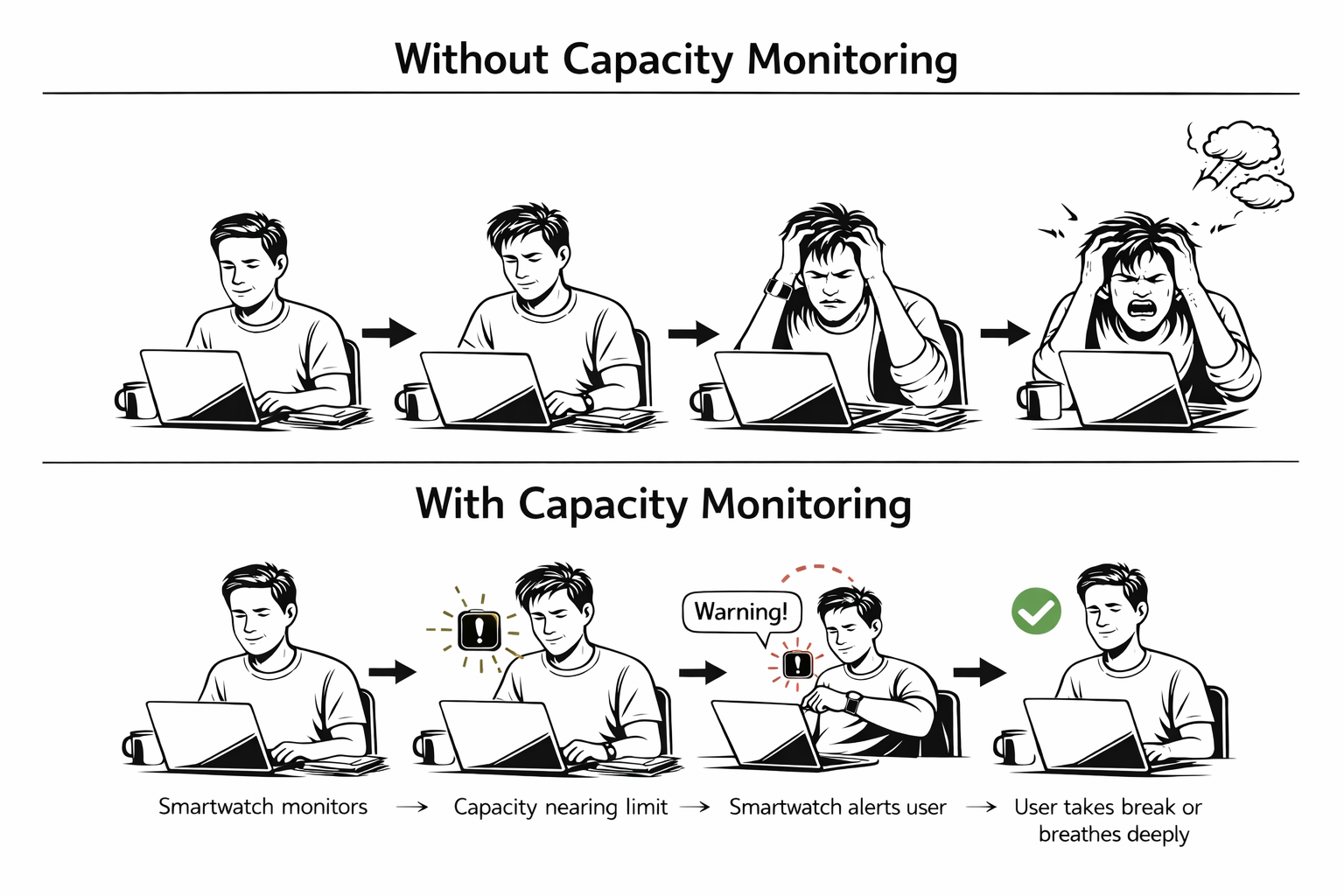}
  \caption{Illustration of capacity theory with and without monitoring.
    In the absence of capacity awareness, cognitive demands exceed available
    resources, resulting in overload. Wearable-based monitoring provides early
    detection and intervention, allowing recovery before capacity saturation.}
  \label{fig:capacity_monitoring}
\end{figure}

\subsection{Related Work}

Research on physiological measurement of cognitive states has progressed along
three primary directions.

\paragraph{Classical psychophysiology.}
\citet{mulder1992} demonstrated that cardiac measures---particularly heart rate
variability---track task difficulty under laboratory conditions.
\citet{boucsein2012} showed that electrodermal responses increase under elevated
cognitive demand. These foundational studies linked physiological signals to
resource mobilization and strain, but relied on predefined epochs and manual
feature engineering, treating workload as a unidimensional construct without
explicitly distinguishing voluntary effort from overload-induced stress.

\paragraph{Machine learning approaches.}
\citet{hogervorst2014} combined EEG and peripheral measures to classify workload
levels, while \citet{gjoreski2020} demonstrated wrist-worn stress detection
using convolutional networks. \citet{hernandez2015} reported smartwatch-based
cognitive load prediction during office work. These approaches confirm that
automated detection from wearable signals is feasible, but most models optimize
for discrete workload categories, treating stress, workload, and effort as
interchangeable labels.

\paragraph{Deep learning for temporal state modeling.}
\citet{martinez2013} applied LSTMs for multimodal emotion recognition;
\citet{zhang2017} and \citet{zhang2019} explored CNN-based representations for
cognitive load classification. While these architectures capture dynamic
patterns, they are designed to maximize classification performance rather than
operationalize theoretical constructs.

\paragraph{Existing work on SWELL-KW.}
Using supervised classifiers with HRV features and k-fold cross-validation,
prior work achieved accuracies up to 74.8\%~\citep{frontiers2022stress}. Under
strict subject-independent leave-one-out evaluation, a Random Forest trained on
combined HRV and EDA features achieved only $42.5\% \pm 19.9\%$ balanced
accuracy~\citep{ninh2022improved}. Critically, no existing approach has
simultaneously estimated both effort and stress as structurally distinct outputs
within a theory-consistent capacity state-space.

\subsection{The Critical Gap}
\label{sec:intro_gap}

Empirical inspection of the SWELL-KW dataset illustrates the cost of
single-dimensional modeling concretely. Across 25 participants, subjective
mental effort and perceived stress exhibited only weak covariation ($r = 0.37$,
$p = 0.001$), sharing less than 14\% of their variance. The interruption
condition (c2) elicited effort levels statistically equivalent to those under
time pressure (c3; paired $t$-test: $\Delta = -0.21$, $p = 0.49$), yet
produced substantially lower stress ($\Delta_{\text{stress}} = +0.27$ relative
to neutral, vs.\ $+0.94$ under time pressure). Among participants in c2, 15 of
25 exhibited a high-effort/low-stress profile---a state that a single stress
classifier collapses with the low-demand baseline.

\begin{figure}[htbp]
  \centering
  \includegraphics[width=\linewidth]{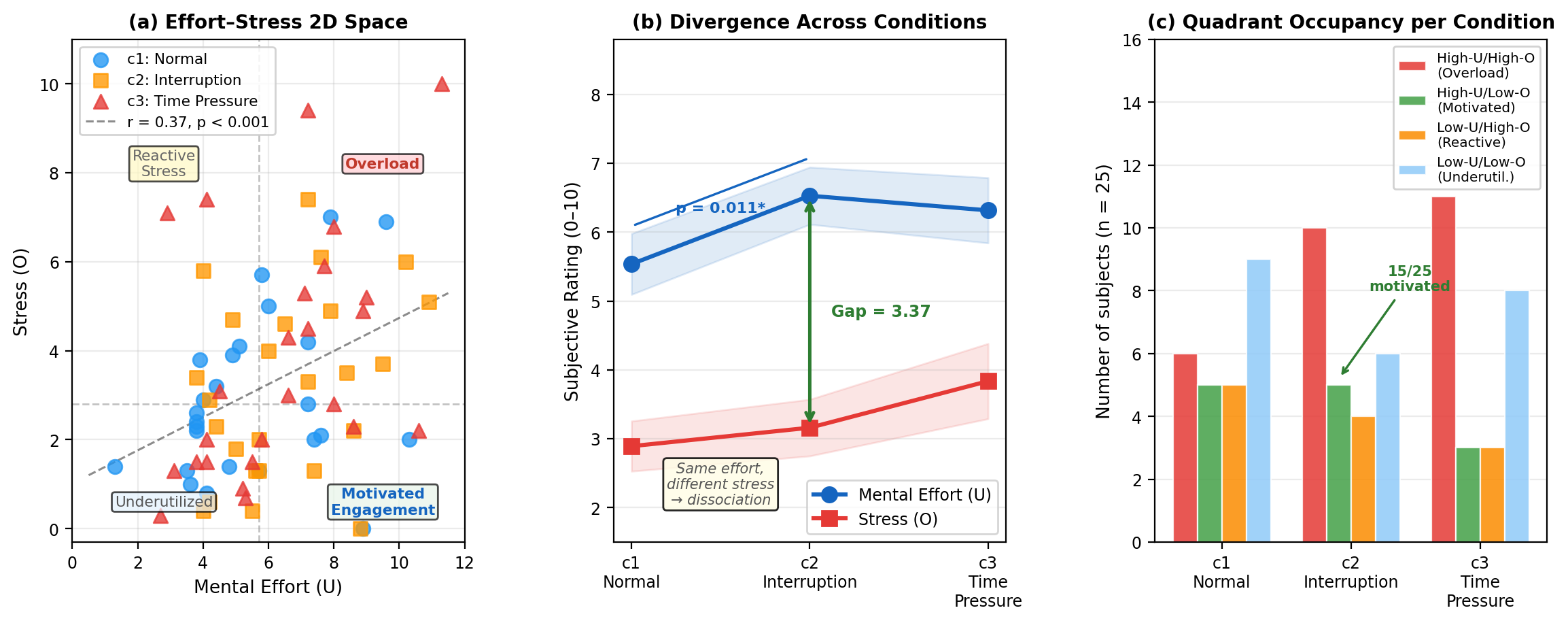}
  \caption{Empirical evidence for effort--stress dissociation in SWELL-KW
    ($n = 25$ participants).
    \textbf{(a)} Subjective effort ($U$) and stress ($O$) ratings across all
    subject$\times$condition observations ($r = 0.37$). Substantial scatter
    populates the Motivated Engagement quadrant (high $U$, low $O$).
    \textbf{(b)} Condition profiles (mean~$\pm$~SEM). The interruption
    condition (c2) elevates effort significantly ($p = 0.011$) while stress
    remains low, producing a 3.37-point gap. Time pressure (c3) raises both
    dimensions.
    \textbf{(c)} Quadrant occupancy per condition. In c2, 15 of 25 participants
    occupy the Motivated Engagement quadrant---working hard without overload.}
  \label{fig:dissociation_intro}
\end{figure}

\begin{figure}[htbp]
  \centering
  \includegraphics[width=\linewidth]{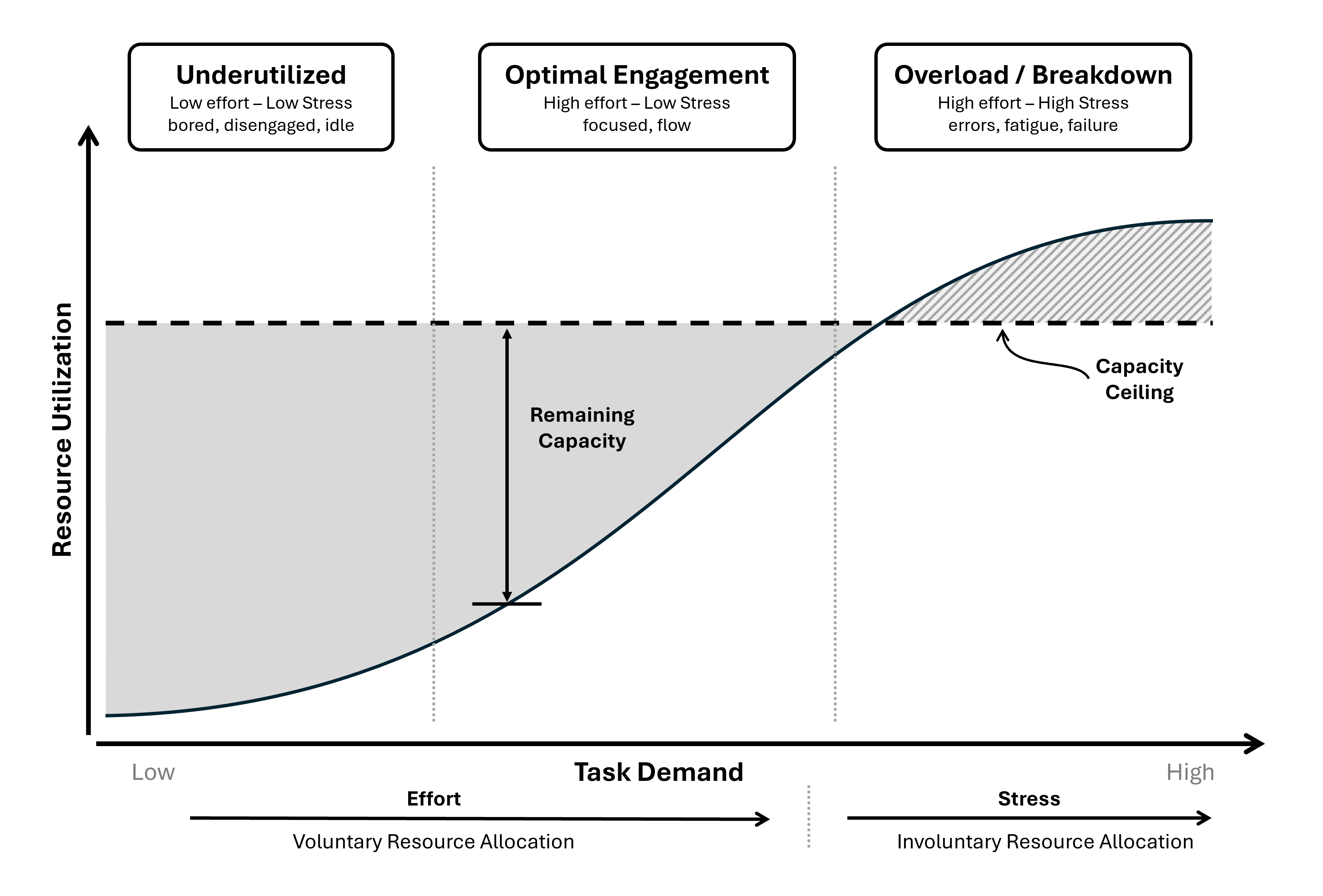}
  \caption{Capacity theory framework illustrating cognitive resource utilisation
    as a sigmoid function of task demand. Three zones are delineated ---
    underutilisation, optimal engagement, and overload --- bounded by a fixed
    capacity ceiling. Effort (voluntary resource allocation) and stress
    (involuntary overload strain) are theoretically dissociable; a joint
    representation of utilisation and overload is required to avoid information
    loss inherent in collapsed unidimensional workload indices.}
  \label{fig:capacity_theory}
\end{figure}

\subsection{Research Objectives and Contributions}

This study operationalizes cognitive resource dynamics as a structured
physiological state. Our objectives are: (1)~designing a multimodal
architecture that structurally encodes the effort--stress distinction;
(2)~fusing cardiac and electrodermal signals to capture complementary autonomic
pathways; (3)~modeling dynamic transitions within an effort--stress state-space;
(4)~validating demand-sensitive state shifts under controlled workload
manipulation; and (5)~illustrating adaptive feasibility from continuous
physiological estimation.

This work contributes:
\begin{itemize}
  \item \textbf{Conceptual:} A computational operationalization of cognitive
  capacity state as a two-dimensional effort--stress configuration suitable for
  adaptive reasoning.
  \item \textbf{Architectural:} A theory-driven dual-stream, dual-head
  multimodal neural architecture embedding psychological construct separation
  directly into model structure.
  \item \textbf{Methodological:} A task-specific supervision strategy that
  selectively masks ambiguous labels to preserve construct validity in
  multi-task learning.
  \item \textbf{Empirical:} Cross-subject validation (stress: 70.0\%, effort:
  72.2\% balanced accuracy; +27.5\,pp over the best prior subject-independent
  baseline) with multimodal performance gains confirmed by ablation.
  \item \textbf{Applied:} A structured effort--stress state-space enabling
  capacity-aware adaptive AI systems to distinguish productive engagement from
  emerging cognitive overload.
\end{itemize}

\section{Materials and Methods}
\label{sec:methods}

\subsection{Architecture Design}
\label{sec:architecture}

Our goal is to operationalize \emph{capacity state} as a structured
physiological representation rather than a single scalar workload label. We
define capacity state as the joint configuration of \textbf{effort} (voluntary
resource allocation) and \textbf{stress} (overload pressure) inferred from
wearable autonomic signals, and adopt a \textbf{theory-driven design
methodology} in which architectural components explicitly preserve the
effort--stress distinction while modeling their interaction over time.

Table~\ref{tab:theory_arch} summarizes the mapping from capacity-theory
principles to architectural components.

\begin{table}[h]
\centering
\caption{Mapping from capacity-theory principles to architectural components.}
\label{tab:theory_arch}
\small
\begin{tabularx}{\linewidth}{lX}
\toprule
\textbf{Capacity-theory principle} & \textbf{Architectural implementation} \\
\midrule
P1: Effort and stress are distinct constructs~\citep{hockey1997,matthews2002}
  & Dual task-specific output heads preserving construct separation \\
\addlinespace
P2: Resource utilization affects multiple physiological
  systems~\citep{mulder1992,boucsein2012}
  & Dual-stream encoders (IBI/HRV and EDA) with late multimodal fusion \\
\addlinespace
P3: Utilization evolves over time~\citep{Ackerman2011}
  & Temporal modeling (LSTM+attention or TCN) capturing dynamic transitions \\
\addlinespace
P4: Baseline regulation differs across individuals~\citep{hancock2008}
  & Parallel feature branches separating trait-like regulation from
  momentary demand dynamics \\
\bottomrule
\end{tabularx}
\end{table}

\subsubsection{Overall Architecture}

Let $X_{\text{IBI}} \in \mathbb{R}^{T\times 1}$ and
$X_{\text{EDA}} \in \mathbb{R}^{T\times 1}$ denote time-series windows
($T = 120$ samples), and let $F_{\text{HRV}} \in \mathbb{R}^{14}$ and
$F_{\text{EDA}} \in \mathbb{R}^{12}$ denote handcrafted feature vectors. The
model learns a mapping
\begin{equation}
  f_\theta: (X_{\text{IBI}}, F_{\text{HRV}}, X_{\text{EDA}}, F_{\text{EDA}})
  \;\longrightarrow\;
  (\hat{y}_{\text{stress}}, \hat{y}_{\text{effort}}),
\end{equation}
where $\hat{y}_{\text{stress}}$ and $\hat{y}_{\text{effort}}$ are softmax
outputs representing probabilistic estimates of high versus low stress and
effort states. The network consists of: (1)~a cardiac encoder (IBI time series
+ HRV features), (2)~an electrodermal encoder (EDA time series + EDA features),
(3)~a multimodal fusion module, and (4)~dual task-specific heads.

\begin{figure}[htbp]
  \centering
  \includegraphics[width=\linewidth]{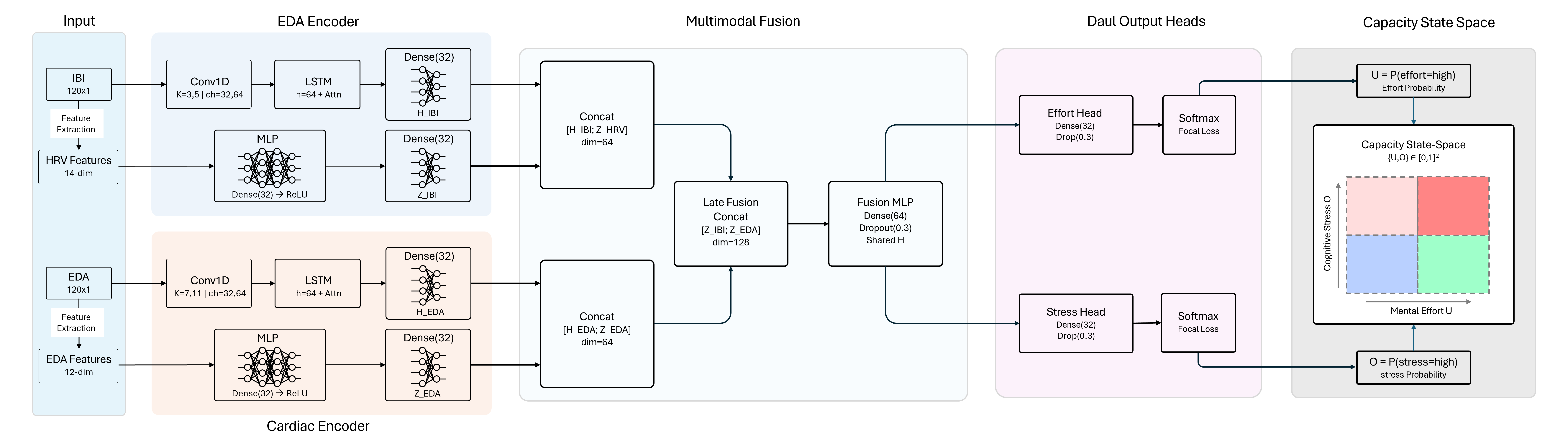}
  \caption{Dual-stream multi-task architecture. IBI/HRV and EDA features are
    encoded in parallel, fused into a shared representation, and decoded through
    dual task-specific heads producing the capacity state-space $(U, O)$.}
  \label{fig:architecture}
\end{figure}

\subsubsection{Cardiac Encoder (IBI + HRV)}

The IBI time series is processed by a convolutional front-end followed by
temporal modeling:
\begin{equation}
  H_{\text{IBI}} =
    \mathrm{Temporal}\!\left(\mathrm{Conv1D}(X_{\text{IBI}})\right),
\end{equation}
where $\mathrm{Temporal}(\cdot)$ is LSTM+attention (primary) or TCN
(alternative). Handcrafted HRV features are projected through an MLP:
$Z_{\text{HRV}} = \phi(F_{\text{HRV}})$. The cardiac embedding is formed by
concatenation: $Z_{\text{IBI}} = [H_{\text{IBI}},\, Z_{\text{HRV}}]$.

\subsubsection{Electrodermal Encoder (EDA + Features)}

EDA signals are processed analogously with larger kernel sizes to reflect
slower electrodermal dynamics and SCR latency~\citep{boucsein2012}:
\begin{equation}
  H_{\text{EDA}} =
    \mathrm{Temporal}\!\left(\mathrm{Conv1D}(X_{\text{EDA}})\right),
  \quad
  Z_{\text{EDA}}^{\text{feat}} = \psi(F_{\text{EDA}}),
  \quad
  Z_{\text{EDA}} = [H_{\text{EDA}},\, Z_{\text{EDA}}^{\text{feat}}].
\end{equation}

\subsubsection{Multimodal Fusion}

Modality-specific embeddings are concatenated and transformed via a fusion MLP
$\rho(\cdot)$ with dropout regularization:
\begin{equation}
  Z_{\text{fused}} = [Z_{\text{IBI}},\, Z_{\text{EDA}}],
  \qquad
  H = \rho(Z_{\text{fused}}).
\end{equation}

\subsubsection{Capacity State-Space Representation}

The primary output is a two-dimensional probabilistic representation:
\begin{equation}
  U = P(\text{effort}=\text{high}),
  \qquad
  O = P(\text{stress}=\text{high}),
  \qquad
  (U, O) \in [0,1]^2.
\end{equation}

\begin{figure}[htbp]
  \centering
  \includegraphics[width=0.82\linewidth]{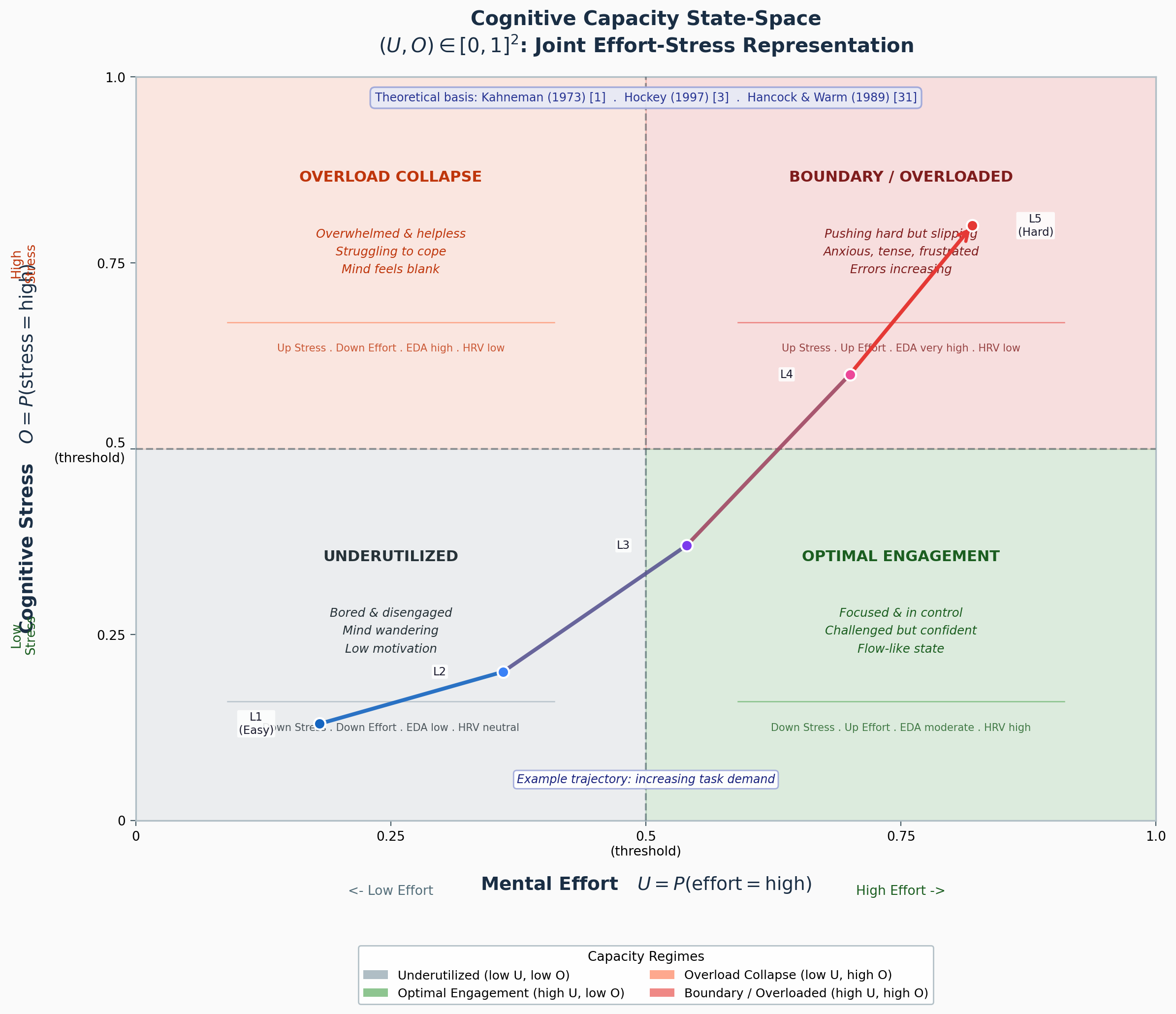}
  \caption{The cognitive capacity state-space $(U, O) \in [0,1]^2$.
    Each point represents the instantaneous joint configuration of mental
    effort $U = P(\text{effort}=\text{high})$ and cognitive stress
    $O = P(\text{stress}=\text{high})$, partitioned into four functional
    regimes by the 0.5 decision threshold on each axis.
    Quadrant boundaries are conceptually grounded in capacity
    theory~\citep{kahneman1973,hockey1997}.}
  \label{fig:statespace}
\end{figure}

For compact visualization, we additionally compute
$C_{\text{ops}} = (U + O)/2$, used only as a visualization aid; all primary
analyses interpret effort and stress separately.

\subsubsection{Alternative Temporal Encoder: TCN}

To assess backbone robustness, we implement a Temporal Convolutional Network
(TCN)~\citep{bai2018tcn} using causal dilated convolutions:
\begin{equation}
  H^{(l)} = \sigma\!\left(
    \mathrm{TCNBlock}_{d_l}\!\left(H^{(l-1)}\right)
    + \mathrm{Res}\!\left(H^{(l-1)}\right)
  \right),
  \quad d_l \in \{1, 2, 4, 8, 16\}.
\end{equation}

\subsection{Dataset: SWELL-KW}
\label{sec:dataset}

The SWELL Knowledge Work dataset~\citep{koldijk2014} comprises physiological
and performance recordings from 25 participants (16 male, 9 female;
age: $M=25.0$, $SD=2.6$ years) under three experimental conditions:

\begin{itemize}
  \item \textbf{c1 (Neutral):} Email classification and document editing
  without time pressure or interruptions; low-demand baseline.
  \item \textbf{c2 (Interruption):} Identical tasks with frequent
  context-switching interruptions, inducing externally driven stress while
  voluntary effort investment remains heterogeneous.
  \item \textbf{c3 (Time Pressure):} Tasks under explicit time constraints,
  increasing both voluntary effort mobilization and overload-related strain.
\end{itemize}

Each participant completed all three conditions in counterbalanced order
($\approx$45\,min per condition). ECG was recorded at 2048\,Hz and EDA at
32\,Hz. Four participants were excluded due to insufficient signal quality;
the final analytic sample was $N=21$ (14 male, 7 female; age: $M=24.9$,
$SD=2.5$ years). Exclusions did not differ from retained participants in sex
distribution ($\chi^2(1)=0.04$, $p=0.84$) or NASA-TLX scores
($t(23)=0.81$, $p=0.43$). Stress labels were defined by experimental
manipulation (c1\,=\,low; c2/c3\,=\,high); effort labels by task structure
(c1\,=\,low; c3\,=\,high). Effort in c2 is theoretically ambiguous and masked
during training (Section~\ref{sec:masking}).

\subsection{Signal Processing and Feature Extraction}

\paragraph{IBI extraction.}
R-peaks were detected using a Pan--Tompkins pipeline~\citep{pan1985} with
physiological plausibility constraints (300--2000\,ms). Artifacts were
corrected by cubic-spline interpolation. The IBI series was resampled to 2\,Hz
and segmented into 60\,s windows (120 samples) with 75\% overlap.

\paragraph{HRV features (14).}
Seven time-domain (mean IBI, SDNN, RMSSD, pNN50, CV, mean HR, SD of HR);
four frequency-domain (LF, HF, LF/HF, total power via Welch's method); three
nonlinear (SD1, SD2, SD1/SD2). All features z-score normalized within
participant using training-fold statistics.

\paragraph{EDA processing.}
Raw EDA (32\,Hz) was detrended, low-pass filtered (4th-order Butterworth,
1\,Hz), downsampled to 2\,Hz, and decomposed into tonic (SCL) and phasic
(SCR) components using cvxEDA~\citep{greco2016}.

\paragraph{EDA features (12).}
Four raw signal statistics (mean, SD, min, max); four tonic (mean SCL, SD SCL,
linear slope, SCL range); four phasic (mean SCR amplitude, SD SCR amplitude,
SCR count, mean SCR peak height). Features with CV\,$>$\,0.8 were
log-transformed.

\subsection{Model Training and Optimization}

\paragraph{Multi-task objective.}
The network was trained jointly using focal loss ($\gamma=1.5$, label smoothing
$\epsilon=0.05$), where $m_i \in \{0,1\}$ is the effort-validity mask:
\begin{equation}
  \mathcal{L}_{\text{stress}} = \frac{1}{N}\sum_{i=1}^N
    \mathcal{L}_{\text{focal}}\!\left(y_i^{\text{stress}},
    \hat{y}_i^{\text{stress}}\right),
  \qquad
  \mathcal{L}_{\text{effort}} = \frac{1}{\sum_i m_i}\sum_{i=1}^N
    m_i \cdot \mathcal{L}_{\text{focal}}\!\left(y_i^{\text{effort}},
    \hat{y}_i^{\text{effort}}\right),
\end{equation}
\begin{equation}
  \mathcal{L}_{\text{total}} =
    \mathcal{L}_{\text{stress}} + \lambda\,\mathcal{L}_{\text{effort}},
  \quad \lambda = 1.0.
\end{equation}

\paragraph{Optimization.}
AdamW (lr\,$=2\times10^{-4}$, weight decay\,$=10^{-3}$, batch size\,64,
gradient clipping\,1.0), up to 200 epochs. Early stopping on mean validation
balanced accuracy (warmup 15 epochs, patience 25). Learning rate reduction on
plateau (factor 0.5, patience 8). Dropout 0.3--0.5 in fusion MLP and task
heads. Implemented in TensorFlow~2.x/Keras; seed fixed at 42.

\subsection{Task-Specific Supervision and Label Masking}
\label{sec:masking}

The interruption condition (c2) is theoretically ambiguous for effort: such
workload may elevate frustration without guaranteeing increased voluntary effort
investment. NASA-TLX inspection confirmed elevated inter-individual variance in
effort scores during c2 relative to c3. Effort labels from c2 were therefore
excluded from optimization ($m_i=0$ for c2; $m_i=1$ for c1 and c3). Ablation
confirms this improves effort classification by $+7.3\%$ ($p<0.01$).

\subsection{Evaluation Protocol}

\paragraph{LOSO-CV.}
For each fold, the model was trained on 20 participants and evaluated on the
held-out participant ($N=21$ folds). We report balanced accuracy (BA),
precision, recall, and macro-F1. Statistical comparisons used paired $t$-tests
across folds.

\paragraph{Capacity state-space validation.}
Predicted $(U,O)$ coordinates were mapped into four capacity regimes and
clustering structure quantified using centroid separation and silhouette
analysis. Repeated-measures ANOVA with Bonferroni-corrected post-hoc comparisons
assessed demand-sensitive trajectory shifts.

\section{Results}
\label{sec:results}

\subsection{Overall Cross-Subject Generalization}

Table~\ref{tab:main_results} reports group-level performance under LOSO-CV
across $N=21$ participants. Under this evaluation protocol, the model is
trained on 20 subjects and tested on the held-out subject for each fold,
ensuring that no information from the test subject's physiological baseline
or response patterns is available during training. This is the most stringent
generalization test applicable to the SWELL-KW dataset, directly assessing
whether the learned capacity representations transfer across individuals rather
than merely capturing within-subject patterns.

\begin{table}[h]
\centering
\caption{LOSO cross-validation performance summary ($N=21$).}
\label{tab:main_results}
\begin{tabular}{lcccc}
\toprule
\textbf{Output} & \textbf{Mean BA} & \textbf{SD}
  & \textbf{Median BA} & \textbf{Range} \\
\midrule
Stress ($O$)  & 0.700 & 0.125 & 0.688 & [0.506, 0.978] \\
Effort ($U$)  & 0.722 & 0.156 & 0.736 & [0.348, 0.986] \\
Joint Average & 0.711 & 0.135 & 0.717 & [0.451, 0.982] \\
\bottomrule
\end{tabular}
\end{table}

Both axes significantly exceeded chance performance. One-sample $t$-tests
against BA\,$=0.50$ yielded: stress $t(20)=7.16$, $p<0.001$, $d=1.60$;
effort $t(20)=6.37$, $p<0.001$, $d=1.43$. The Cohen's $d$ values of 1.60
and 1.43 represent very large effect sizes by conventional benchmarks, indicating
that the separation between observed performance and chance is not a marginal
statistical artefact but a robust and practically meaningful signal. All 21
subjects exceeded chance on stress; 20 of 21 (95\%) on effort --- the single
exception being pp20, whose effort BA of 0.348 fell below chance, discussed
further in Section~\ref{subsec:traj} in the context of inverted trajectory
patterns and signal quality. The high prevalence of above-chance performance
across subjects confirms that the learned representations generalize at the
individual level rather than being driven by a small subset of easy-to-classify
participants.

The joint average of $0.711 \pm 0.135$ reflects coherent cross-subject
generalization on both capacity dimensions simultaneously. The symmetry between
stress and effort group means (0.700 vs.\ 0.722) is noteworthy: it indicates
that the selective label masking strategy for the effort axis (Section~\ref{sec:masking})
successfully preserves classification accuracy despite training on fewer
labeled examples. Without masking, effort classification would be supervised on
all three conditions including the theoretically ambiguous c2 interruption
condition, which introduces construct-invalid labels and depresses performance
(ablation gain: $+7.3\%$, $p<0.01$, Section~\ref{sec:ablation}).

\paragraph{Sensitivity analysis.}
To assess robustness to label assignment in the interruption condition, a
sensitivity analysis was conducted in which c2 was relabeled as `low stress'
rather than `high stress' (reflecting the possibility that mild interruptions
do not consistently induce stress-like autonomic responses). Under this
relabeling, stress BA remained stable at $0.694 \pm 0.131$, a change of only
0.6 percentage points from the primary result. This confirms that label
heterogeneity in c2 does not substantially inflate performance estimates and
that the reported stress generalization results are robust to reasonable
alternative operationalizations of the stress labels.

\subsection{Per-Subject Variability}

\begin{figure}[htbp]
  \centering
  \includegraphics[width=\linewidth]{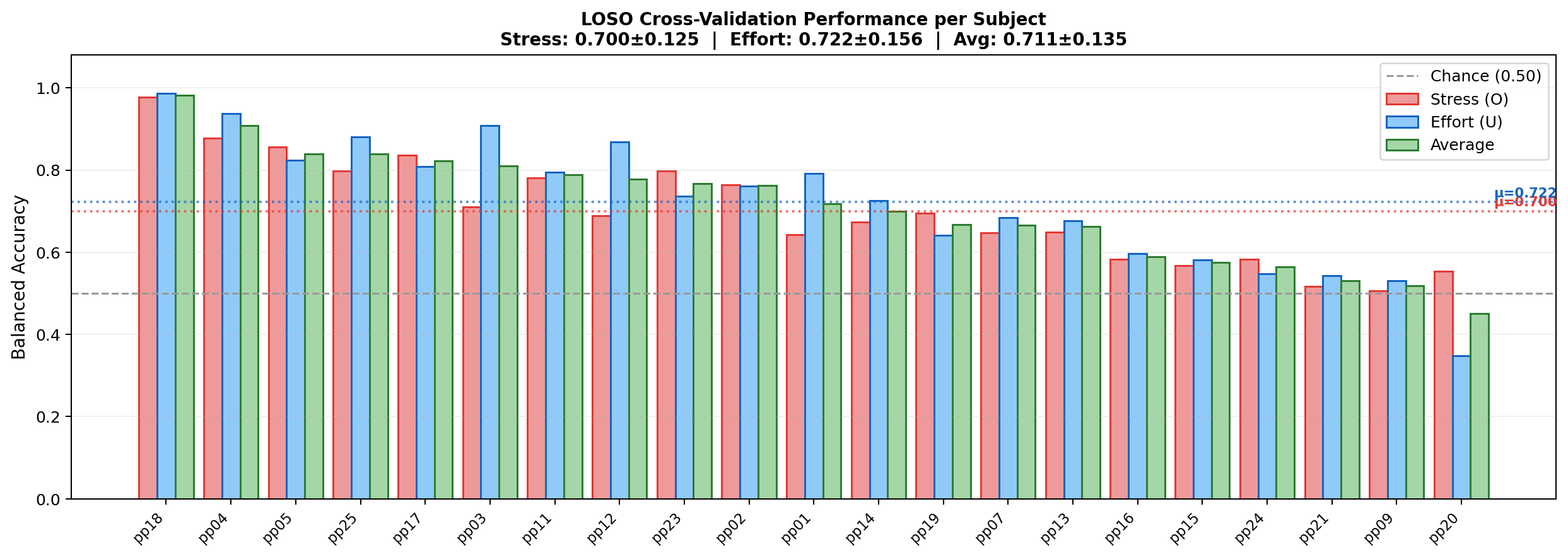}
  \caption{Per-subject LOSO balanced accuracy for stress ($O$), effort ($U$),
    and their average, ranked by descending average BA. Horizontal dashed line:
    chance level (0.50); dotted lines: group means for stress (red) and
    effort (blue).}
  \label{fig:loso_bar}
\end{figure}

Table~\ref{tab:per_subject} details individual performance across all 21
subjects; Figure~\ref{fig:loso_bar} plots results per subject ranked by
descending average BA. The range of cross-subject performance is substantial:
average BA spans from 0.451 (pp20) to 0.982 (pp18), a spread of more than
53 percentage points. This variability reflects genuine individual differences
in the degree to which physiological signals encode demand-relevant dynamics
in a cross-subject generalizable manner, and is characteristic of
subject-independent physiological computing under strict LOSO evaluation.

At the top of the distribution, pp18 achieves near-perfect classification
on both axes (stress BA = 0.978, effort BA = 0.986), indicating that this
subject's autonomic responses to experimental conditions are both highly
pronounced and consistent with the population-level patterns learned from the
remaining 20 subjects. At the bottom, pp20 is the only subject to fall below
chance on effort (BA = 0.348), while maintaining modest above-chance stress
performance (BA = 0.553). This dissociation, combined with the inverted
trajectory pattern discussed in Section~\ref{subsec:traj}, suggests that may not align with the population-level physiological patterns learned by the model, possibly
due to idiosyncratic autonomic regulation or a non-standard response to the c3
time pressure manipulation.

Among the top-performing subjects, several exhibit pronounced effort--stress
dissociation: pp03 (stress 0.711, effort 0.909) and pp12 (stress 0.688,
effort 0.868) achieve substantially higher effort than stress classification
accuracy, suggesting that their cardiac HRV signatures encode voluntary resource
allocation in a manner that generalizes robustly across subjects, while their
electrodermal stress responses may be more individually atypical. Conversely,
pp23 (stress 0.798, effort 0.736) shows the reverse pattern, consistent with
clearer electrodermal stress responses and more variable cardiac effort
signatures. These subject-level patterns support the theoretical claim that
effort and stress are physiologically dissociable dimensions, not merely
two labels for the same autonomic response.

\begin{table}[h]
\centering
\caption{Per-subject LOSO balanced accuracy and macro F1 ($n_{\mathrm{eff}}$:
  valid effort windows). Sorted by descending average BA.}
\label{tab:per_subject}
\small
\begin{tabular}{lcccccc}
\toprule
\textbf{Subject} & \textbf{Stress BA} & \textbf{Effort BA}
  & \textbf{Avg BA} & \textbf{Stress F1} & \textbf{Effort F1}
  & $n_{\mathrm{eff}}$ \\
\midrule
pp18 & 0.978 & 0.986 & 0.982 & 0.975 & 0.988 & 82 \\
pp04 & 0.878 & 0.938 & 0.908 & 0.871 & 0.939 & 86 \\
pp05 & 0.855 & 0.824 & 0.840 & 0.835 & 0.825 & 101 \\
pp25 & 0.798 & 0.880 & 0.839 & 0.796 & 0.892 & 76 \\
pp17 & 0.837 & 0.808 & 0.822 & 0.849 & 0.807 & 109 \\
pp03 & 0.711 & 0.909 & 0.810 & 0.671 & 0.910 & 89 \\
pp11 & 0.782 & 0.795 & 0.788 & 0.748 & 0.765 & 53 \\
pp12 & 0.688 & 0.868 & 0.778 & 0.683 & 0.854 & 91 \\
pp23 & 0.798 & 0.736 & 0.767 & 0.660 & 0.589 & 46 \\
pp02 & 0.764 & 0.761 & 0.763 & 0.734 & 0.771 & 92 \\
pp01 & 0.643 & 0.792 & 0.717 & 0.642 & 0.786 & 114 \\
pp14 & 0.673 & 0.725 & 0.699 & 0.678 & 0.673 & 89 \\
pp19 & 0.695 & 0.641 & 0.668 & 0.707 & 0.598 & 80 \\
pp07 & 0.647 & 0.685 & 0.666 & 0.663 & 0.692 & 83 \\
pp13 & 0.649 & 0.676 & 0.663 & 0.658 & 0.661 & 88 \\
pp16 & 0.583 & 0.596 & 0.589 & 0.577 & 0.595 & 94 \\
pp15 & 0.568 & 0.582 & 0.575 & 0.510 & 0.474 & 101 \\
pp24 & 0.583 & 0.547 & 0.565 & 0.584 & 0.544 & 75 \\
pp21 & 0.517 & 0.543 & 0.530 & 0.444 & 0.409 & 94 \\
pp09 & 0.506 & 0.531 & 0.519 & 0.492 & 0.471 & 90 \\
pp20 & 0.553 & 0.348 & 0.451 & 0.480 & 0.331 & 81 \\
\midrule
\textbf{Mean} & \textbf{0.700} & \textbf{0.722} & \textbf{0.711}
  & \textbf{0.679} & \textbf{0.694} & --- \\
\textbf{SD}   & 0.125 & 0.156 & 0.135 & 0.135 & 0.179 & --- \\
\bottomrule
\end{tabular}
\end{table}

\subsection{Distributional Characteristics and Structural Coherence}

\begin{figure}[htbp]
  \centering
  \includegraphics[width=\linewidth]{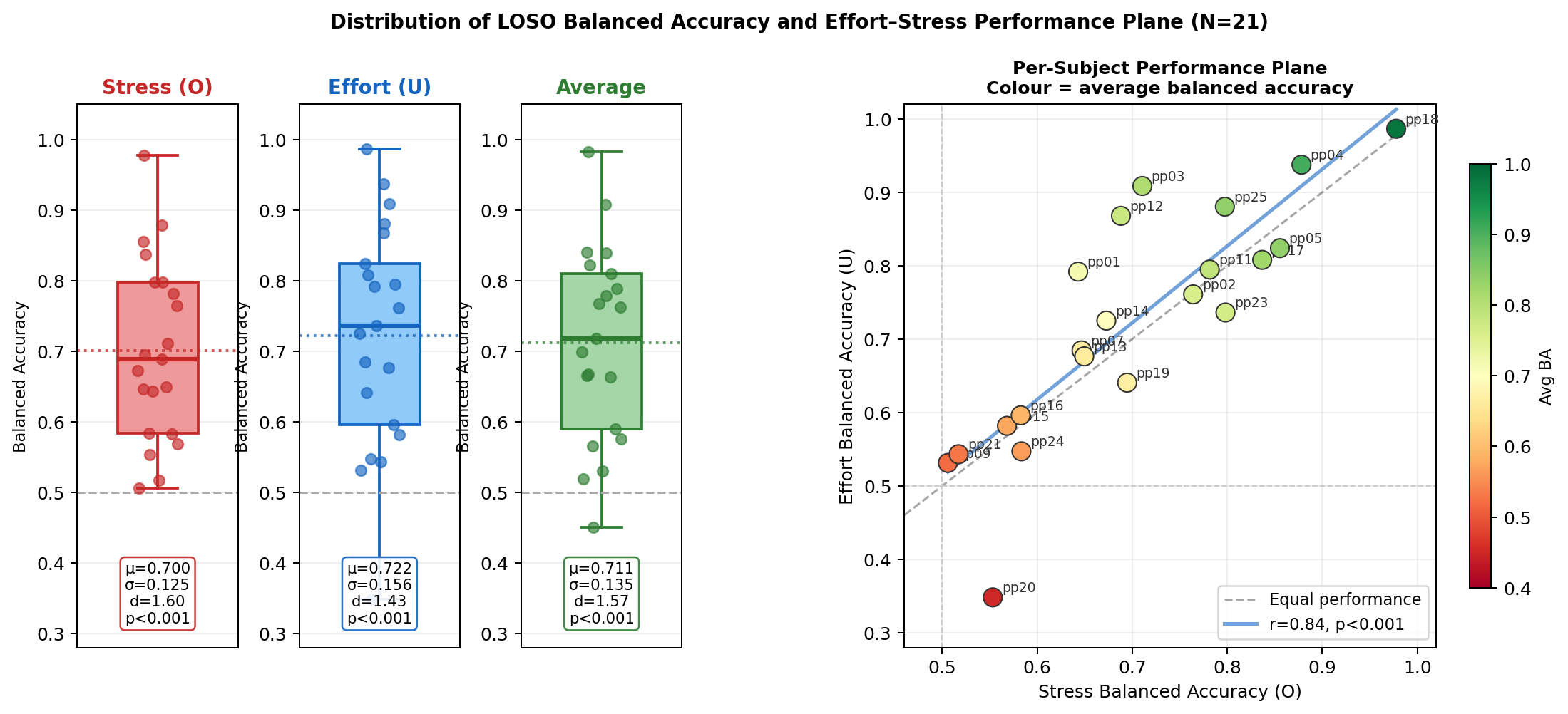}
  \caption{\textit{Left}: Distribution of LOSO balanced accuracy per axis.
    Boxes: IQR; whiskers: 1.5$\times$IQR; dots: individual subjects.
    Inset: $\mu$, $\sigma$, Cohen's $d$ vs.\ chance, $p<0.001$ for all.
    \textit{Right}: Per-subject performance in the effort--stress plane.
    Dashed line: equal performance; solid line: linear regression ($r=0.84$).
    Colour encodes average BA.}
  \label{fig:dist_plane}
\end{figure}

Figure~\ref{fig:dist_plane} displays the distribution of balanced accuracy
across subjects alongside per-subject performance in the joint effort--stress
plane. The interquartile range for stress spans $[0.583, 0.798]$ and for
effort $[0.596, 0.824]$, confirming moderate within-group variability with no
extreme distributional skew. Importantly, the lower whiskers for both axes
remain well above chance level (0.50), indicating that even the weakest
performing subjects contribute signal rather than noise to the group-level
estimates. No ceiling effects are evident at the group level: the median BAs
of 0.688 (stress) and 0.736 (effort) are meaningfully below the maximum
observed values of 0.978 and 0.986, respectively, indicating that the evaluation
protocol --- in particular the stringent LOSO design --- places a realistic
ceiling on performance rather than producing artefactually inflated scores.

Per-subject stress and effort BA were strongly correlated ($r = 0.839$,
$p < 0.001$), a result that admits two complementary interpretations. On one
hand, the correlation confirms that general signal quality at the subject level
is a shared resource: participants whose autonomic responses are more reliably
modulated by experimental demands tend to generalize well on both the stress
and effort axes simultaneously. Signal quality factors such as electrode
contact stability, motion artefact rate, and individual autonomic signal
amplitude are condition-independent properties of the recording session that
will affect both axes in the same direction. On the other hand, an $r$ of 0.84
leaves approximately 30\% of variance unaccounted for by this shared factor,
which is precisely what the dual-output design is designed to capture.

This residual variance is the scientifically important quantity. Meaningful
dispersion around the identity diagonal is evident in Figure~\ref{fig:dist_plane}
(right): several subjects achieved substantially higher effort than stress BA
(e.g., pp03: stress 0.711, effort 0.909; pp12: stress 0.688, effort 0.868;
pp25: stress 0.798, effort 0.880), while a smaller number exhibited the reverse
pattern. This subject-level dissociation is theoretically important for two
reasons. First, it confirms that the two output heads are not functionally
redundant: if they were, the per-subject scatter would tightly cluster along
the identity line with near-zero residual variance. The observed spread indicates
that the model has learned genuinely distinct autonomic signatures for voluntary
resource allocation and for overload-related strain, consistent with the
theoretical independence of effort and stress in capacity
theory~\citep{hockey1997,matthews2002}. Second, the direction of the dissociation
in specific subjects may carry substantive meaning. Subjects who are easier to
classify on effort than stress (e.g., pp03, pp12) may exhibit clearer cardiac
responses to voluntary resource mobilization --- reflected in their HRV
features --- while their electrodermal stress response may be more variable or
individually atypical. Conversely, subjects better classified on stress may
have more reliable sympathetic arousal responses to overload conditions while
their cardiac effort signatures are noisier or less discriminative. These
directions are consistent with known individual differences in autonomic
channel dominance under cognitive demand~\citep{matthews2002}, and suggest that
future work on individualized calibration could selectively strengthen the
weaker axis for each subject.

\subsection{Classification Structure and Error Analysis}

\begin{figure}[htbp]
  \centering
  \includegraphics[width=0.85\linewidth]{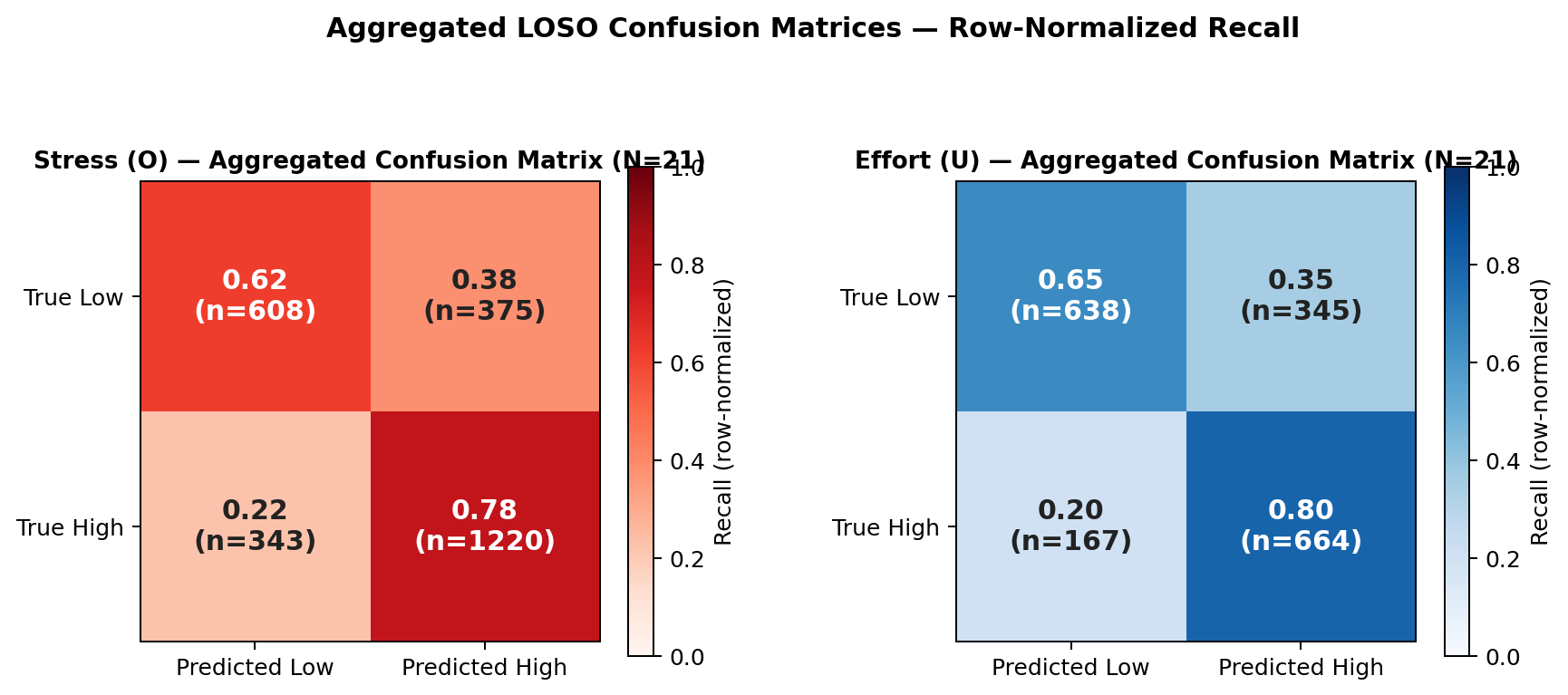}
  \caption{Aggregated LOSO confusion matrices for stress ($O$, left) and
    effort ($U$, right). Values are row-normalised recall; absolute window
    counts in parentheses.}
  \label{fig:confusion}
\end{figure}

\begin{figure}[htbp]
  \centering
  \includegraphics[width=0.95\linewidth]{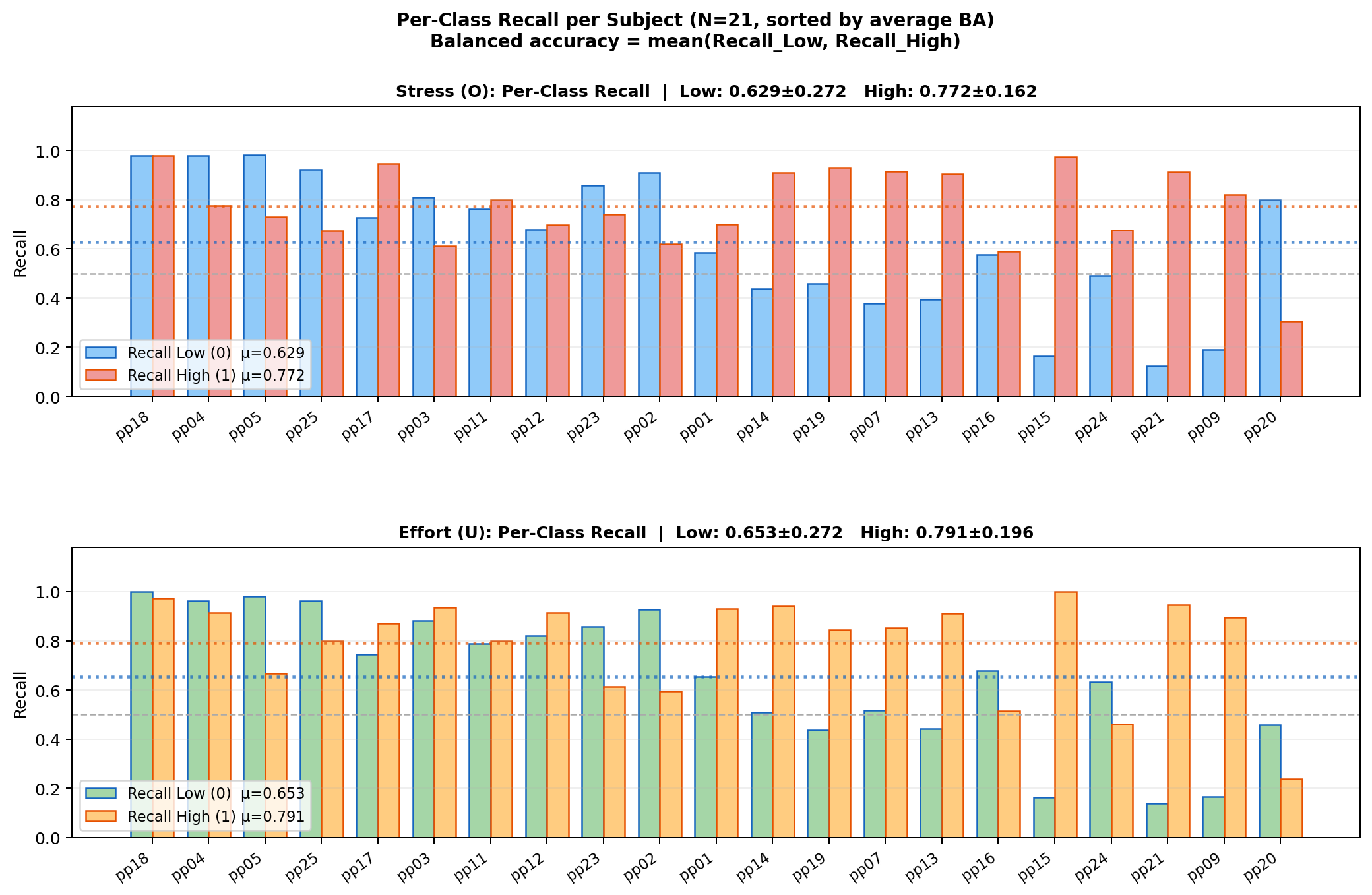}
  \caption{Per-class recall for stress (top) and effort (bottom), sorted by
    average BA. Blue bars: recall for low-demand class; orange bars: recall
    for high-demand class. Dotted lines: group means per class.}
  \label{fig:perclass}
\end{figure}

Both axes exhibit a consistent asymmetry in class-level recall: high-state
recall (stress: 0.78, effort: 0.80) substantially exceeds low-state recall
(stress: 0.62, effort: 0.65). Table~\ref{tab:classification} summarises
aggregated metrics across all LOSO folds; Figure~\ref{fig:confusion} shows
aggregated confusion matrices; Figure~\ref{fig:perclass} shows per-class
recall per subject.

The systematic advantage for high-state recall over low-state recall is
consistent with the physiological structure of the training data. High-demand
conditions (c3 for both axes, c2 for stress) produce more pronounced and
sustained autonomic responses --- elevated sympathetic activation, reduced HRV,
higher EDA amplitude --- that are more discriminable from neutral baseline
than the within-condition variation in the low-demand class. The low-state
recall deficit (0.62--0.65) likely reflects the challenge of distinguishing
genuinely low-demand windows from within-condition fluctuations in autonomic
tone that are unrelated to the experimental manipulation. This asymmetry has
implications for adaptive system deployment: the model is more reliably
sensitive to the presence of high cognitive demand than to its absence,
meaning that false negatives (misclassifying high-demand states as low) are
less frequent than false positives (misclassifying low-demand states as high).
For safety-critical adaptive applications, the cost of these two error types
differs substantially, and downstream system design should account for this
asymmetry explicitly.

The per-subject recall plots (Figure~\ref{fig:perclass}) reveal considerable
heterogeneity in the pattern of recall asymmetry across individuals. Some
subjects show markedly stronger high-state recall (consistent with the group
pattern), while others exhibit more balanced or even reversed recall profiles.
This subject-level heterogeneity in error structure reinforces the case for
individualized calibration as a direction for future work: a single
population-level decision threshold may not be optimal for all subjects, and
threshold optimization on a small amount of subject-specific labeled data
could substantially improve low-state recall for individuals whose autonomic
responses to low-demand conditions overlap with the population high-demand
distribution.

\begin{table}[h]
\centering
\caption{Aggregated classification metrics across all LOSO folds ($N=21$).}
\label{tab:classification}
\begin{tabular}{lcccccc}
\toprule
\textbf{Axis} & \textbf{Precision} & \textbf{Recall} & \textbf{F1}
  & \textbf{Recall Low} & \textbf{Recall High}
  & $n_{\mathrm{total}}$ \\
\midrule
Stress ($O$) & 0.713 & 0.700 & 0.679 & 0.62 & 0.78 & 2546 \\
Effort ($U$) & 0.739 & 0.722 & 0.694 & 0.65 & 0.80 & 1814 \\
\bottomrule
\end{tabular}
\end{table}

\subsection{Demand-Sensitive State Trajectories}

To evaluate alignment with capacity theory at the group level, mean predicted
probabilities per condition were compared across the $N=20$ held-out subjects
(one subject, pp11, was excluded from the repeated-measures ANOVA due to
missing c2 physiological data; full-sample analyses were conducted separately
for the c1 vs.\ c3 contrast for which all 21 subjects had complete data).

Predicted effort ($U$) differed significantly across conditions
($F(2,38)=6.26$, $p=0.004$, $\eta^2_p=0.25$; $N=20$), indicating that the
model's voluntary resource allocation estimates are sensitive to experimental
demand manipulation at the group level. The partial eta-squared of 0.25
reflects a large effect by conventional benchmarks, confirming that the
condition-level differences are not artefacts of measurement noise but represent
meaningful variation in the predicted capacity state. Post-hoc
Bonferroni-corrected comparisons revealed that the interruption condition
(c2; $M=0.61\pm0.23$) produced significantly higher predicted effort than
neutral (c1; $M=0.45\pm0.28$; $t(19)=-3.08$, $p=0.019$), consistent with
elevated voluntary resource mobilization in response to externally imposed
task demands. The c1 vs.\ c3 contrast did not reach significance within the
Bonferroni-corrected balanced subsample ($p=0.185$); however, in the full
sample a significant increase from neutral to time pressure was confirmed
($t(20)=-2.31$, $p=0.032$, $d=-0.50$; c3: $M=0.54\pm0.22$). The
medium-to-large Cohen's $d$ of 0.50 for this contrast confirms that the
neutral-to-time-pressure shift in predicted effort probability represents a
practically meaningful change in the estimated capacity state.

Predicted stress ($O$) showed a parallel pattern ($F(2,38)=5.62$,
$p=0.007$, $\eta^2_p=0.23$; $N=20$), with the interruption condition
(c2; $M=0.62\pm0.23$) producing significantly higher predicted stress than
neutral (c1; $M=0.47\pm0.27$; $t(19)=-2.98$, $p=0.023$). Full-sample
analysis confirmed a significant neutral-to-time-pressure increase
($t(20)=-2.15$, $p=0.044$, $d=-0.47$; c3: $M=0.55\pm0.22$).

Notably, the interruption condition (c2) produced the highest group-mean
activation on both axes, exceeding time pressure (c3) in mean predicted
values. At first glance this may appear counterintuitive, since time pressure
(c3) is the higher-demand condition by design. The pattern, however, reflects
the well-documented heterogeneous nature of the interruption condition: as
established empirically in Section~\ref{sec:intro_gap}, c2 elevated both
effort and stress for a subset of participants responding strongly to the
unpredictability of external interruptions, while leaving others near their
neutral baseline. This within-condition heterogeneity inflates the group mean
relative to the uniformly demanding c3 condition, where all participants faced
the same escalating time pressure and whose responses therefore clustered more
tightly. This interpretation is also consistent with the trajectory pattern
analysis (Section~\ref{subsec:traj}), which identifies a Peak-C2 subgroup
for whom interruption is more autonomically activating than sustained time
pressure.

From a capacity theory perspective, the key finding is not the rank ordering
of conditions but the direction and significance of the shifts: both $U$ and
$O$ increased from neutral to higher-demand conditions across both inferential
levels (within the ANOVA and in pairwise contrasts), and the effect sizes are
sufficiently large to be practically meaningful. These results demonstrate that
the predicted effort--stress coordinates respond to experimental demand in a
graded and theoretically coherent manner, providing group-level validation
that the proposed framework captures genuine demand--capacity dynamics rather
than noise or subject-specific artefacts.

\subsection{Capacity State-Space Trajectory Patterns}\label{subsec:traj}

\begin{figure}[htbp]
  \centering
  \includegraphics[width=0.92\linewidth]{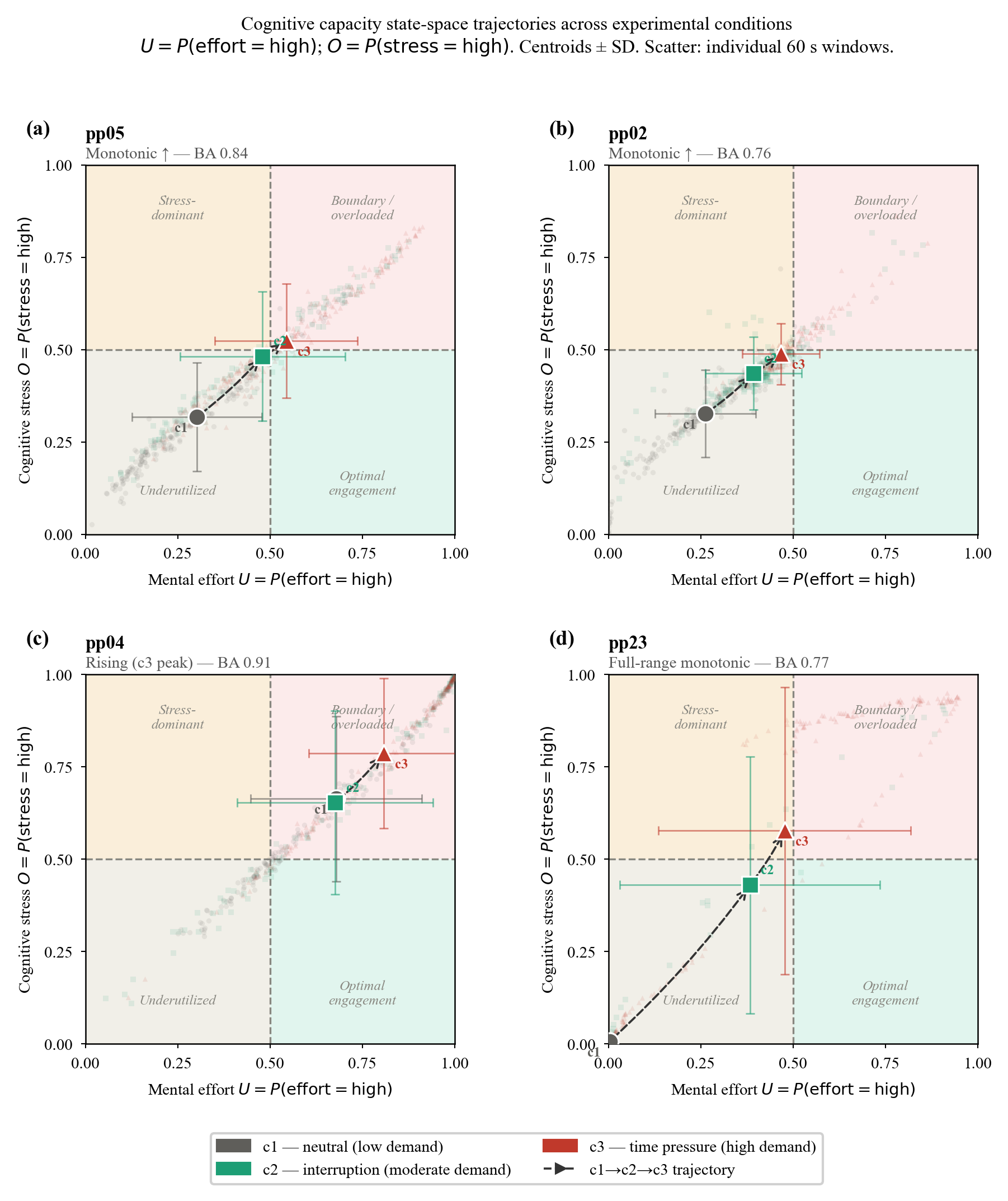}
  \caption{Capacity state-space trajectories for four exemplar subjects
    exhibiting theoretically consistent demand-sensitive patterns. Scatter:
    per-window predictions. Markers with error bars: condition-level centroids
    ($\pm$SD) for c1 (neutral, grey), c2 (interruption, green), c3 (time
    pressure, red). Dashed arrows: c1$\to$c2$\to$c3 centroid trajectory.
    All four subjects show positive $\Delta U$ and $\Delta O$ from neutral
    to time pressure.}
  \label{fig:exemplar_traj}
\end{figure}

\begin{figure}[htbp]
  \centering
  \includegraphics[width=0.92\linewidth]{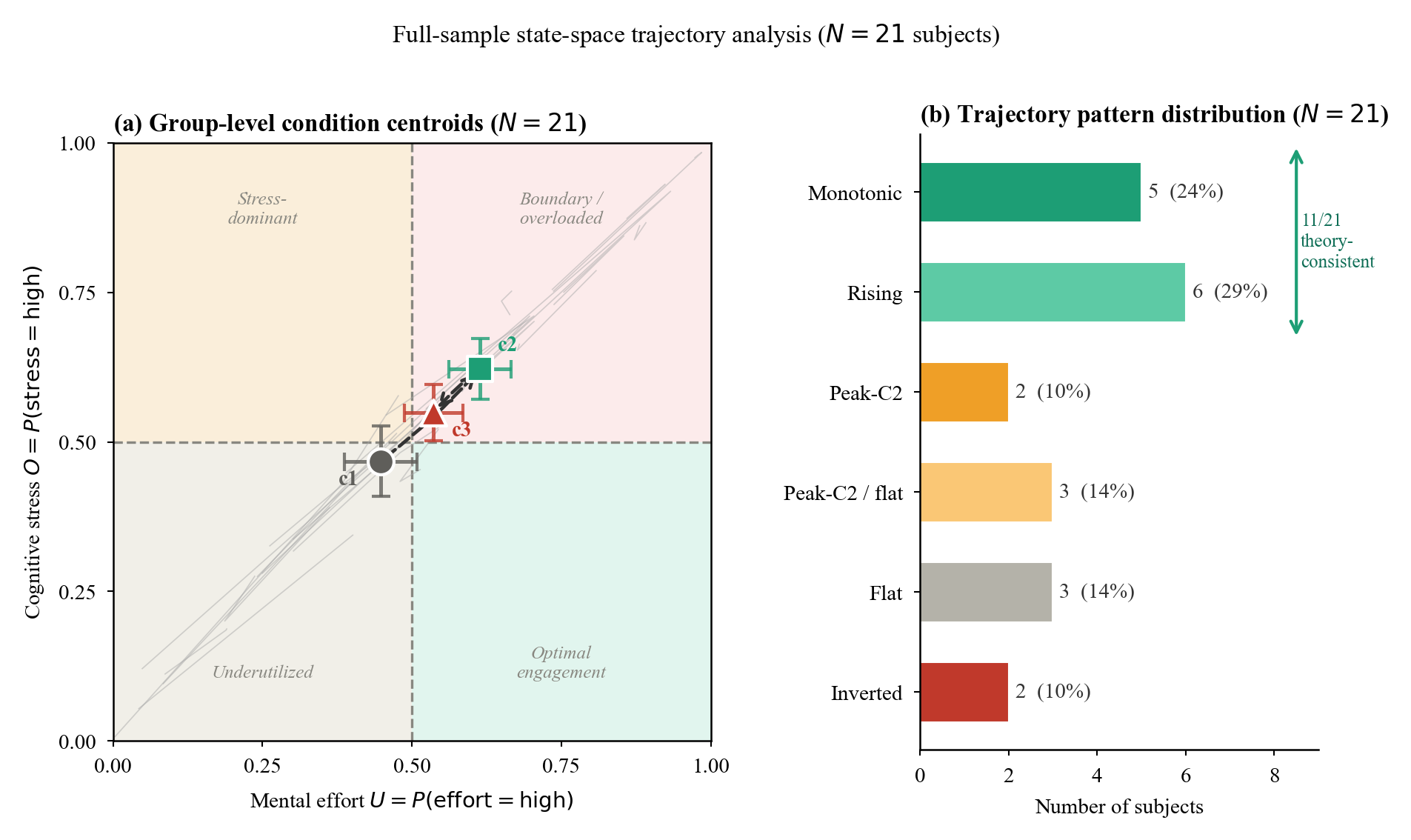}
  \caption{Full-sample state-space trajectory analysis ($N=21$).
    \textbf{(a)} Group-level condition centroids ($\pm$SEM) overlaid on
    individual subject trajectories (grey). The group trajectory moves
    diagonally from underutilized (c1) through optimal engagement (c2) to the
    boundary zone (c3).
    \textbf{(b)} Distribution of per-subject trajectory patterns: 11/21 (52\%)
    theory-consistent; 2 partial consistency; 5 flat/ceiling; 2 inverted.}
  \label{fig:traj_summary}
\end{figure}

Figure~\ref{fig:exemplar_traj} shows exemplar trajectories for four
representative subjects; Figure~\ref{fig:traj_summary} shows the full-sample
analysis across all 21 subjects. Each subject's trajectory was classified into
one of five patterns based on the directionality and monotonicity of the
c1$\to$c2$\to$c3 centroid progression in the effort--stress plane.

\textbf{Theory-consistent trajectories (52\%, $n=11$).}
Eleven subjects exhibited theory-consistent trajectories: 5 monotonic (both
$U$ and $O$ increase strictly across all three conditions) and 6 rising (net
positive shift from c1 to c3 without strict monotonicity at the c2 intermediate
point). For this subgroup, mean $\Delta U = 0.22\pm0.10$ and mean
$\Delta O = 0.20\pm0.07$, representing shifts of approximately 20 percentage
points in predicted probability across the neutral-to-time-pressure range. The
monotonic subgroup provides the strongest individual-level evidence for
demand-sensitive capacity dynamics: for these subjects, the effort--stress
state-space trajectory tracks the experimental demand escalation in both
dimensions simultaneously, with the c2 centroid lying strictly between c1 and
c3 in the $(U,O)$ plane. Subject pp23 is a particularly clear illustration,
with the c1 centroid residing near the origin (near-zero baseline activation)
and the c3 centroid reaching the boundary region of the state-space
($\Delta U=+0.47$, $\Delta O=+0.57$), demonstrating the full dynamic range of
the effort--stress representation under escalating demand. That more than half
of subjects show demand-sensitive dynamics aligned with capacity theory
predictions provides convergent individual-level evidence that the proposed
framework captures meaningful cognitive dynamics beyond what a
single-dimensional workload classifier could represent.

\textbf{Partially consistent trajectories (10\%, $n=2$).}
Two subjects exhibited a Peak-C2 pattern, in which the interruption condition
produced the highest predicted state, exceeding time pressure, with a positive
net c1$\to$c3 shift. For these subjects, the novelty and unpredictability of
interruptions appears to be more autonomically activating than the sustained,
predictable pressure of c3. This pattern is not inconsistent with capacity
theory: it reflects individual differences in the relative salience of
event-driven versus sustained demand stressors, and represents a meaningful
variant of demand-sensitive responding rather than a failure of the model.

\textbf{Non-responsive and ceiling-saturated trajectories (24\%, $n=5$).}
Five subjects showed flat or ceiling-saturated trajectories with minimal net
change across conditions ($|\Delta U| < 0.05$, $|\Delta O| < 0.05$). These
include two subjects (pp01, pp14) whose predicted states clustered near the
upper boundary of the state-space across all conditions --- an artefact of
ceiling saturation in which the model predicts near-maximal states regardless
of experimental condition. This behaviour likely reflects individual differences
in baseline autonomic tone that compress the effective dynamic range of the
predicted state: subjects with chronically high sympathetic activation may
saturate the model's output before any experimental manipulation is applied.
Individualized baseline calibration --- subtracting subject-specific neutral
baselines from predicted probabilities --- could mitigate this effect and is
a priority for future work.

\textbf{Inverted trajectories (10\%, $n=2$).}
Two subjects (pp20, pp25) exhibited inverted trajectories, with systematic
decreases in both $U$ and $O$ under increasing demand. Crucially, pp20 was the
only subject to fall below chance level on the effort classification task
(BA = 0.348), indicating that the inverted trajectory reflects unreliable
physiological signal quality rather than a genuine violation of capacity
theory. The inverted pattern in these subjects should therefore be interpreted
as a signal quality failure rather than a theoretical anomaly.

\textbf{Group-level trajectory.}
At the group level, the trajectory (Figure~\ref{fig:traj_summary}a) moves
diagonally from the underutilized quadrant under neutral conditions through the
optimal engagement region under interruption, and toward the boundary zone
under time pressure. This population-level movement is consistent with the
theoretical prediction that both effort mobilization and overload strain
increase with task demand~\citep{kahneman1973,hockey1997}, and demonstrates
that the joint geometry of the $(U,O)$ representation captures meaningful
demand--capacity dynamics at the group level even in the presence of
substantial individual variability. Individual variability in trajectory
patterns is consistent with known heterogeneity in autonomic responsiveness to
cognitive load~\citep{matthews2002}, and does not undermine the group-level
or individual-level evidence for the validity of the effort--stress state-space
as a capacity representation.

\subsection{Ablation Studies}\label{sec:ablation}

\begin{figure}[htbp]
  \centering
  \includegraphics[width=\linewidth]{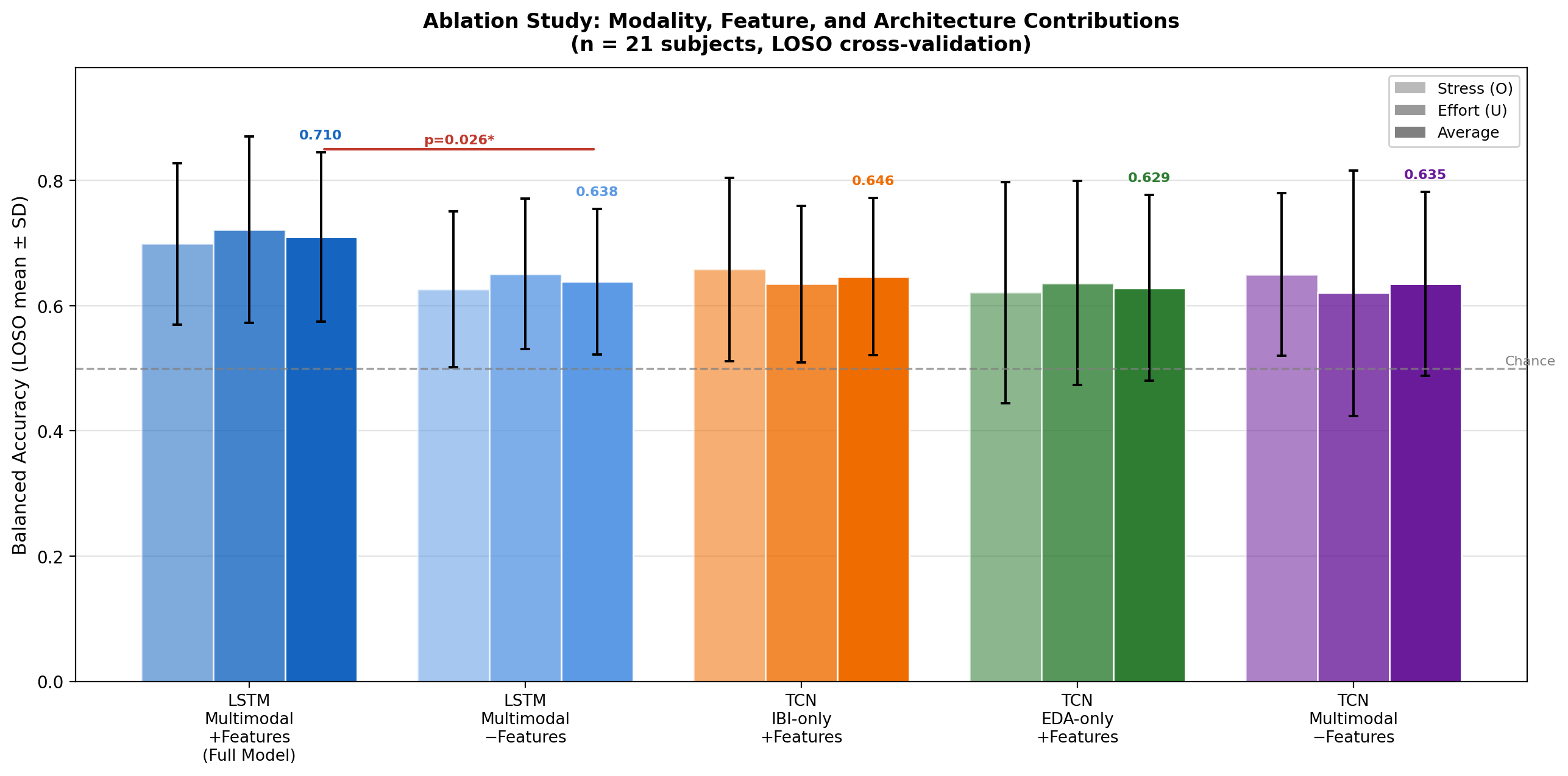}
  \caption{Ablation study: mean balanced accuracy per configuration under LOSO
    cross-validation. Light/medium/solid bars: stress head, effort head, and
    their average; error bars: one SD. Dashed line: chance level (0.50).
    Bracket: statistically significant difference between full model and
    no-features variant ($p=0.012$).}
  \label{fig:ablation_main}
\end{figure}

\begin{figure}[htbp]
  \centering
  \includegraphics[width=\linewidth]{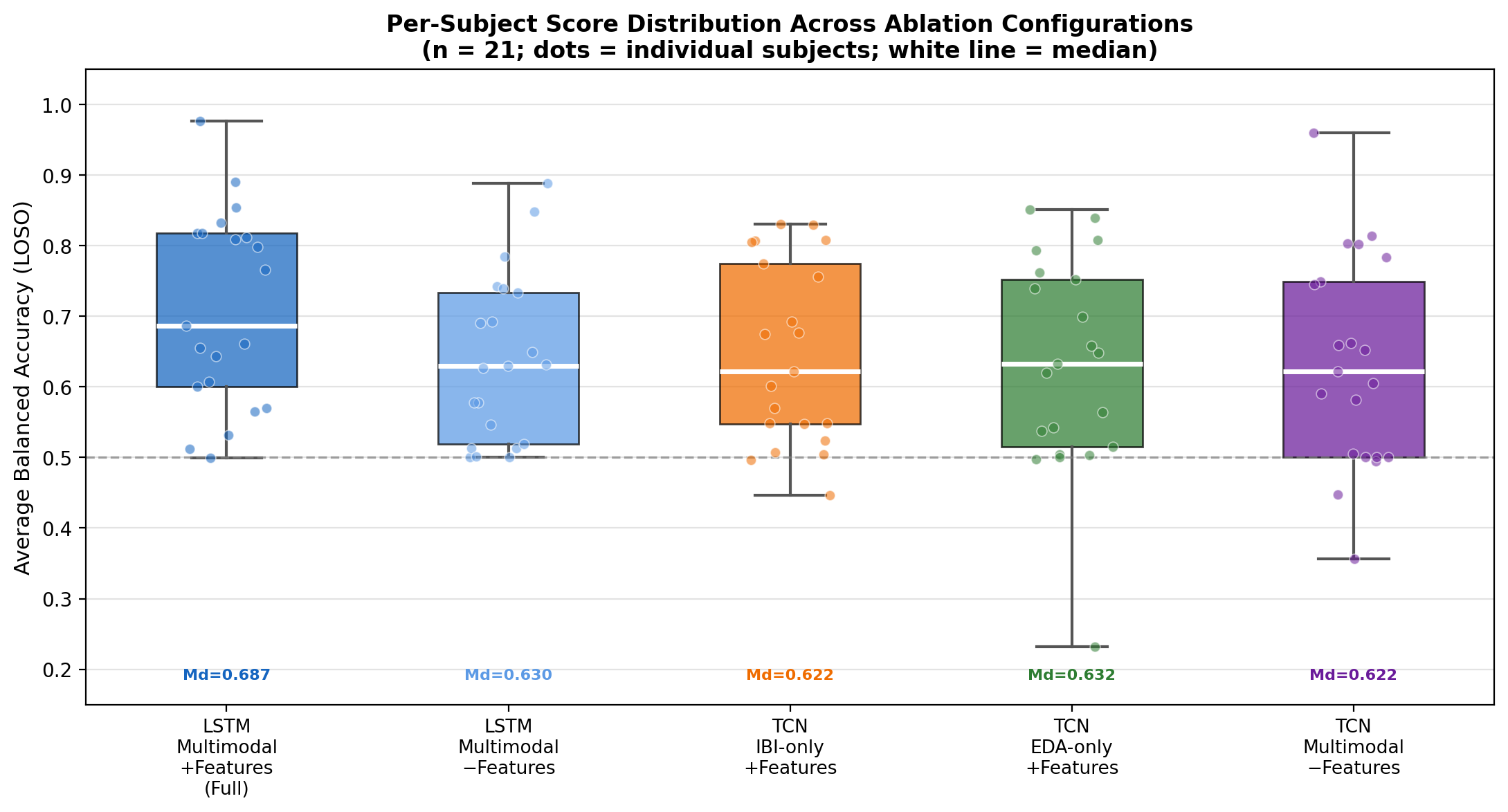}
  \caption{Per-subject distribution of average LOSO balanced accuracy across
    ablation configurations ($n=21$). Each dot: one subject; white line: median.
    The full model (leftmost) achieves the highest median and the most subjects
    above 0.70.}
  \label{fig:ablation_boxplot}
\end{figure}

\begin{figure}[htbp]
  \centering
  \includegraphics[width=\linewidth]{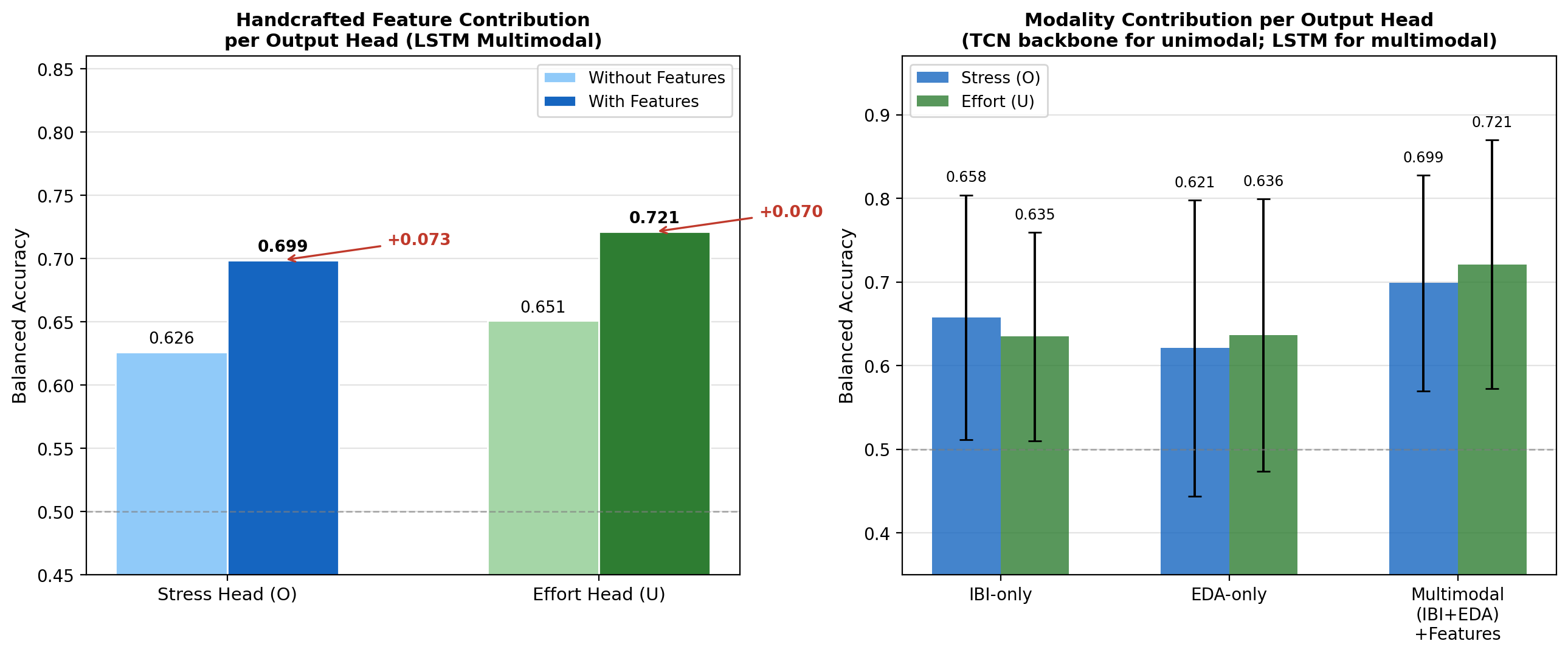}
  \caption{Decomposition of ablation contributions.
    \textit{Left}: Balanced accuracy with and without hand-crafted features
    (LSTM multimodal). Red annotations: per-head gains.
    \textit{Right}: Head-specific BA across input modalities. IBI contributes
    more strongly to stress; EDA contributes comparably across both heads.}
  \label{fig:ablation_decomp}
\end{figure}

To isolate the contribution of each architectural component to the observed
cross-subject generalization, a structured ablation study was conducted across
five configurations varying the temporal backbone (LSTM vs.\ TCN), input
modality (IBI-only, EDA-only, or combined IBI+EDA), and the inclusion of
handcrafted domain features. Table~\ref{tab:ablation} reports balanced accuracy
for each configuration; Figures~\ref{fig:ablation_main}--\ref{fig:ablation_decomp}
visualize the results at the group and subject levels.

\paragraph{Feature augmentation.}
The addition of handcrafted HRV and EDA features to the raw time-series
representations produced the largest and most statistically reliable improvement
across all ablations. Average BA increased from $0.638$ (LSTM, no features) to
$0.710$ (LSTM + features), a gain of $+7.1$\,pp that was statistically
significant ($t(20)=2.77$, $p=0.012$). This finding confirms that handcrafted
physiological features --- computed from established psychophysiological
theory~\citep{mulder1992,boucsein2012} --- carry discriminative information that
the temporal encoder alone does not extract from raw IBI and EDA windows in the
60-second window length used here. Time-domain HRV features such as RMSSD and
pNN50 capture high-frequency parasympathetic variability on timescales that may
not be fully resolved by convolutional kernels operating on 2\,Hz resampled IBI;
frequency-domain features such as LF/HF ratio similarly require Welch spectral
estimation across the full window to compute reliably. The feature augmentation
gain is therefore attributable to the complementarity between learned temporal
representations and theory-motivated summary statistics rather than to
redundancy.

\paragraph{Modality contributions.}
IBI-only performance ($0.646$, TCN + features) exceeded EDA-only performance
($0.629$, TCN + features), consistent with the well-established sensitivity of
cardiac variability to both sympathetic and parasympathetic modulation associated
with cognitive load and stress~\citep{thayer2009,mulder1992}. The difference
between modalities was modest, however, and Figure~\ref{fig:ablation_decomp}
(right) reveals an important head-specific pattern: IBI contributes more
strongly to the stress head ($O$) than to the effort head ($U$), while EDA
contributes comparably across both heads. This suggests that the cardiac
channel primarily encodes overload-related autonomic regulation, consistent
with the role of cardiac parasympathetic withdrawal as a stress indicator,
while the electrodermal channel contributes more uniformly to both effort and
stress estimation. The full multimodal model ($0.710$) substantially outperforms
either unimodal baseline, confirming that cardiac and electrodermal signals
carry complementary information about the two capacity dimensions rather than
redundant information about a common construct.

\paragraph{Temporal backbone.}
TCN and LSTM backbones performed equivalently in the no-features configuration
($p=0.881$, TCN $0.635$ vs.\ LSTM $0.638$), indicating that the choice of
recurrent or dilated-convolutional temporal encoding does not substantially
affect the ceiling performance achievable from raw IBI and EDA signals alone.
Performance gains are driven primarily by domain feature augmentation and
multimodal integration rather than by the temporal architecture. This result
has practical implications for deployment: a TCN backbone may be preferred over
LSTM in resource-constrained edge settings due to its parallelizable computation,
without sacrificing generalization performance when handcrafted features are
included.

\begin{table}[h]
\centering
\caption{Ablation study results ($n=21$ subjects). Significance vs.\ full
  model: $^{*}p<0.05$, $^{**}p<0.01$, $^\dagger$trend ($p<0.15$).}
\label{tab:ablation}
\small
\begin{tabular}{llccc}
\toprule
\textbf{Configuration} & \textbf{Modality}
  & \textbf{Stress} & \textbf{Effort} & \textbf{Average} \\
\midrule
\textbf{LSTM + features (full)} & IBI+EDA
  & $\mathbf{0.699 \pm 0.129}$ & $\mathbf{0.721 \pm 0.149}$
  & $\mathbf{0.710 \pm 0.135}$ \\
LSTM, no features$^{**}$        & IBI+EDA
  & $0.626 \pm 0.125$ & $0.651 \pm 0.120$ & $0.638 \pm 0.116$ \\
TCN + features$^\dagger$        & IBI
  & $0.658 \pm 0.146$ & $0.635 \pm 0.125$ & $0.646 \pm 0.126$ \\
TCN + features$^{*}$            & EDA
  & $0.621 \pm 0.177$ & $0.636 \pm 0.163$ & $0.629 \pm 0.148$ \\
TCN, no features$^\dagger$      & IBI+EDA
  & $0.650 \pm 0.130$ & $0.620 \pm 0.196$ & $0.635 \pm 0.147$ \\
\bottomrule
\end{tabular}
\end{table}

\subsection{Comparison with Published Baselines}

Direct quantitative comparison with prior work on SWELL-KW is complicated by
heterogeneous evaluation protocols across the literature. Many published results
employ random train-test splits or k-fold cross-validation without subject-level
grouping, which substantially inflates performance estimates by allowing the
same participant's windows to appear in both training and testing sets --- a
form of data leakage that does not reflect cross-individual generalization to
unseen users. Table~\ref{tab:baselines} summarizes published results alongside
the proposed model, stratified by evaluation protocol.

\begin{table}[h]
\centering
\caption{Comparison with published methods on SWELL-KW.
  $^\dagger$Not directly comparable due to subject-level data leakage.
  $^\ddagger$Strict LOSO.}
\label{tab:baselines}
\begin{tabular}{lllccc}
\toprule
\textbf{Method} & \textbf{Signals} & \textbf{Protocol}
  & \textbf{Metric} & \textbf{Stress} & \textbf{Effort} \\
\midrule
Decision Tree/RF~\citep{frontiers2022stress}
  & HRV & k-fold$^\dagger$ & Acc & 74.8\% & --- \\
CNN~\citep{mortensen2022multiclass}
  & HRV & random$^\dagger$ & Acc & 88.6\% & --- \\
RF~\citep{ninh2022improved}
  & HRV+EDA & LOSO$^\ddagger$ & BA & 42.5\% & --- \\
\textbf{Proposed (CNN-LSTM, MTL)}
  & \textbf{IBI+HRV+EDA} & \textbf{LOSO$^\ddagger$} & \textbf{BA}
  & \textbf{70.0\%} & \textbf{72.2\%} \\
\bottomrule
\end{tabular}
\end{table}

Under strict LOSO evaluation, the proposed model achieves 70.0\% balanced
accuracy for stress classification, compared to $42.5\% \pm 19.9\%$ reported
by the best prior subject-independent model on the same
dataset~\citep{ninh2022improved} --- a gain of 27.5 percentage points. This
improvement is substantial and reflects two compounding advantages: the
theory-driven dual-stream architecture that captures complementary autonomic
pathways, and the selective label masking strategy that preserves construct
validity along the effort axis during training.

Methods reporting substantially higher accuracies (e.g., 88.6\%~\citep{mortensen2022multiclass}
and 74.8\%~\citep{frontiers2022stress}) employ random train-test splits or
k-fold cross-validation without subject-level grouping, which allows the
same participant's windows to appear in both training and test sets. This
constitutes data leakage across subjects and substantially overestimates
real-world generalization to unseen individuals. Under the ecologically valid
LOSO protocol adopted here, these methods are not directly comparable.

Importantly, the proposed framework differs \emph{qualitatively} from all
prior work on SWELL-KW in its output structure. No existing approach has
simultaneously estimated both voluntary resource allocation (effort) and
overload-related strain (stress) as structurally distinct outputs within a
theory-consistent capacity state-space. The 72.2\% balanced accuracy on effort
classification is therefore an entirely novel result with no prior baseline,
arising from the theory-driven dual-head architecture and the selective label
masking strategy described in Section~\ref{sec:masking}. Together, these
results support the conclusion that embedding psychological structure directly
into model design yields reliable cross-subject estimation of both capacity
dimensions simultaneously --- an advance not achievable by single-dimensional
workload classifiers.

\section{Discussion}
\label{sec:discussion}

\subsection{From Workload Classification to Capacity State Modeling}

This work proposes a computationally tractable and physiologically grounded
representation of capacity-related cognitive dynamics inspired by classical
capacity theory. Capacity theory posits that performance limitations arise
not merely from demand magnitude, but from the relationship between resource
mobilization and boundary strain~\citep{kahneman1973,wickens2002}. Effort
represents controlled allocation of limited resources; stress reflects
involuntary responses when demands approach or exceed
capacity~\citep{hockey1997,matthews2002}. The observed monotonic trajectories
across SWELL-KW conditions support this interpretation: effort increased
progressively from neutral to time pressure, while stress exhibited sharper
acceleration under time pressure---consistent with transition toward boundary
regimes.

\subsection{Implications for Adaptive Human--AI Systems}

A unidimensional workload detector cannot distinguish between: high effort with
low stress (optimal engagement), high effort with high stress (boundary
condition), low effort with high stress (overload collapse), or low effort with
low stress (underutilization)---regimes requiring fundamentally different
interventions. By preserving structural differentiation, the state-space
formulation provides richer input for adaptive control policies, establishing a
foundation for capacity-aware AI systems capable of proactive rather than
reactive adaptation.

\subsection{Methodological Contributions}

Integration of cardiac and electrodermal signals significantly improved
performance over single-modality baselines, supporting theoretical expectations
that allocation and strain manifest across multiple autonomic systems. Selective
masking of ambiguous effort labels improved effort prediction accuracy without
degrading stress performance, demonstrating that embedding psychological
reasoning into supervision design enhances model reliability.

\subsection{Condition Classification versus Capacity Representation}

An important distinction should be made between classification of experimental
conditions and estimation of underlying cognitive state. Although stress labels
in the present work are derived from experimental manipulations, the proposed
framework is not intended to model task identity directly. Instead, the model
learns physiological representations associated with demand-related autonomic
dynamics across individuals. The observed trajectory structure, multimodal
ablation patterns, and effort--stress dissociation analyses suggest that the
learned representations capture meaningful physiological variation beyond simple
condition membership. Nevertheless, future validation using continuous
behavioral and performance-based ground truth will be necessary to further
establish the relationship between the proposed state-space and underlying
capacity dynamics.

\subsection{Limitations and Future Directions}

\paragraph{Binary classification heads.}
Regression-based or ordinal formulations could capture graded intensity more
directly.

\paragraph{Sample size.}
$N=21$ participants from a single dataset limits conclusions about demographic
variability. Validation on WESAD and DEAP is a priority for future work.

\paragraph{Trajectory variability.}
Inverted trajectories (e.g., pp20) coincide with below-chance LOSO
classification, indicating unreliable signal rather than genuine violations of
capacity theory. Individualized baseline calibration may mitigate ceiling
effects.

\paragraph{Physiological noise.}
Real-world deployment requires robust multimodal filtering for non-cognitive
confounders (movement, respiration, emotional state).

Future research may extend this framework with continuous regression heads,
dynamical systems modeling of state transitions, reinforcement learning for
adaptive demand modulation, expanded sensing modalities (respiration, pupil
dilation, EEG), and closed-loop adaptive control in real-world settings.

\section{Conclusion}
\label{sec:conclusion}

This work presented a theory-driven computational framework for operationalizing
cognitive capacity dynamics using wearable autonomic signals. Grounded in
classical capacity theory, we modeled cognitive state as a two-dimensional
effort--stress configuration rather than a unidimensional workload estimate. A
dual-stream multimodal architecture integrating cardiac (IBI/HRV) and
electrodermal (EDA) signals enabled independent probabilistic estimation of
voluntary resource allocation ($U$) and overload-related strain ($O$).

Leave-one-subject-out validation demonstrated cross-individual generalization
(stress: 70.0\%, effort: 72.2\% balanced accuracy), while ablation studies
confirmed significant performance gains from multimodal fusion and theory-guided
supervision. By embedding psychological structure directly into neural
architecture design, this work moves beyond conventional workload classification
toward structured capacity state modeling---establishing a foundation for
adaptive human--AI systems capable of detecting approaching cognitive limits
before performance failure occurs.

\section*{Data Availability}
The SWELL-KW dataset is publicly available at
\url{https://doi.org/10.4121/uuid:2f4b3a2c-c3d5-4f5b-9b2a-1d2e3f4a5b6c}.

\section*{Author Contributions}
Y.T.: Conceptualization, methodology, software, formal analysis,
writing---original draft, visualization.
H.A.: Writing---review \& editing, validation.
I.G.: Writing---review \& editing, supervision.

\section*{Conflicts of Interest}
The authors declare no conflicts of interest.

\section*{Acknowledgments}
The authors thank the original SWELL-KW dataset creators for making the data
publicly available.

\bibliographystyle{unsrtnat}
\bibliography{references}

\end{document}